\documentclass[10pt,twocolumn,letterpaper]{article}

\usepackage{cvpr}
\usepackage{times}
\usepackage{epsfig}
\usepackage{graphicx}
\usepackage{amsmath}
\usepackage{amssymb}
\usepackage{mathtools}
\usepackage{booktabs}
\usepackage{bbm}
\usepackage{algorithm}
\usepackage{algpseudocode}
\usepackage{eqnarray}
\usepackage{array}
\usepackage{relsize}
\usepackage[caption=false]{subfig}
\usepackage{array}
\usepackage{dblfloatfix}
\usepackage{mynotation}

\usepackage[pagebackref=true,breaklinks=true,letterpaper=true,colorlinks,bookmarks=false]{hyperref}

\cvprfinalcopy 

\def\cvprPaperID{****} 

\newcolumntype{C}[1]{>{\centering\let\newline\\\arraybackslash\hspace{0pt}}m{#1}}

\newcommand{\parsection}[1]{\noindent\textbf{#1:} }

\begin{document}

\title{Probabilistic Regression for Visual Tracking}

\newcommand{\asep}{\hspace{6mm}}
\newcommand{\aand}{\hspace{5mm}}
\author{Martin Danelljan \aand Luc Van Gool \aand Radu Timofte\vspace{2mm}\\
Computer Vision Lab, D-ITET, ETH Z\"urich, Switzerland
}

\maketitle

\begin{abstract}
	Visual tracking is fundamentally the problem of regressing the state of the target in each video frame. While significant progress has been achieved, trackers are still prone to failures and inaccuracies. It is therefore crucial to represent the uncertainty in the target estimation. Although current prominent paradigms rely on estimating a state-dependent confidence score, this value lacks a clear probabilistic interpretation, complicating its use.

In this work, we therefore propose a probabilistic regression formulation and apply it to tracking. Our network predicts the conditional probability density of the target state given an input image. Crucially, our formulation is capable of modeling label noise stemming from inaccurate annotations and ambiguities in the task. The regression network is trained by minimizing the Kullback-Leibler divergence. When applied for tracking, our formulation not only allows a probabilistic representation of the output, but also substantially improves the performance. Our tracker sets a new state-of-the-art on six datasets, achieving $59.8\%$ AUC on LaSOT and $75.8\%$ Success on TrackingNet.
The code and models are available at \mbox{\url{https://github.com/visionml/pytracking}}.

\end{abstract}


\section{Introduction}
\label{sec:introduction}

Visual object tracking is the task of estimating the state of a target object in each frame of a video sequence. Most commonly, the state is represented as a bounding box encapsulating the target. Different flavors of the problem arise from  the type of given prior information about the scenario, such as object class~\cite{tracking-bells} or static camera~\cite{tracking-background-model}. In its most general form, however, virtually no prior knowledge is assumed and the initial state of the target is given during inference. This imposes severe challenges, since the method must learn a model of the target during tracking itself.

Along with a myriad of other computer vision tasks, such as object detection~\cite{IOUNet,law2018cornernet,Ren2015FasterRT}, pose estimation~\cite{cao2017realtime,sun2019deep,xiao2018simple}, and keypoint detection~\cite{simonCVPR2017,convposeCVPR2016}, visual tracking can fundamentally be formulated as a regression problem. In this general view, the goal is thus to learn a model, typically a deep neural network, capable of predicting the target state in each frame. While current and past approaches apply a variety of techniques to address this problem, most of the successful methods have a crucial aspect in common. Namely, that the task of regressing the target state $y^*$ in a frame $x$ is achieved by learning to predict a \emph{confidence} value $s(y,x)$ for any given state $y$. The target state is then estimated by \emph{maximizing} the predicted confidence $y^* = \argmax_y s(y,x)$.

\begin{figure}[!t]
	\centering%
	\newcommand{\wid}{0.33\columnwidth}%
	\includegraphics*[trim = 0 0 0 0, width = \wid]{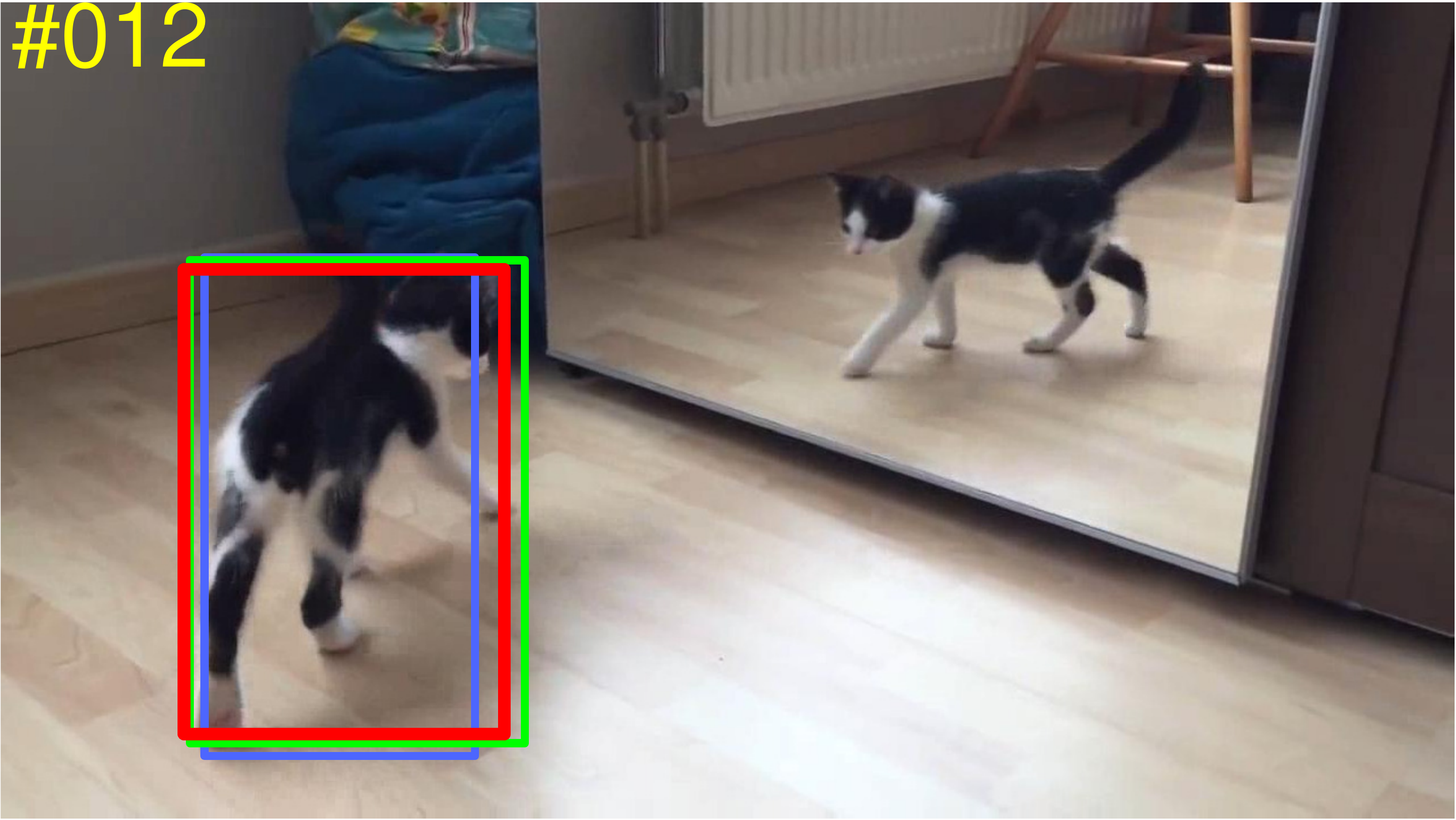}%
	\includegraphics*[trim = 0 0 0 0, width = \wid]{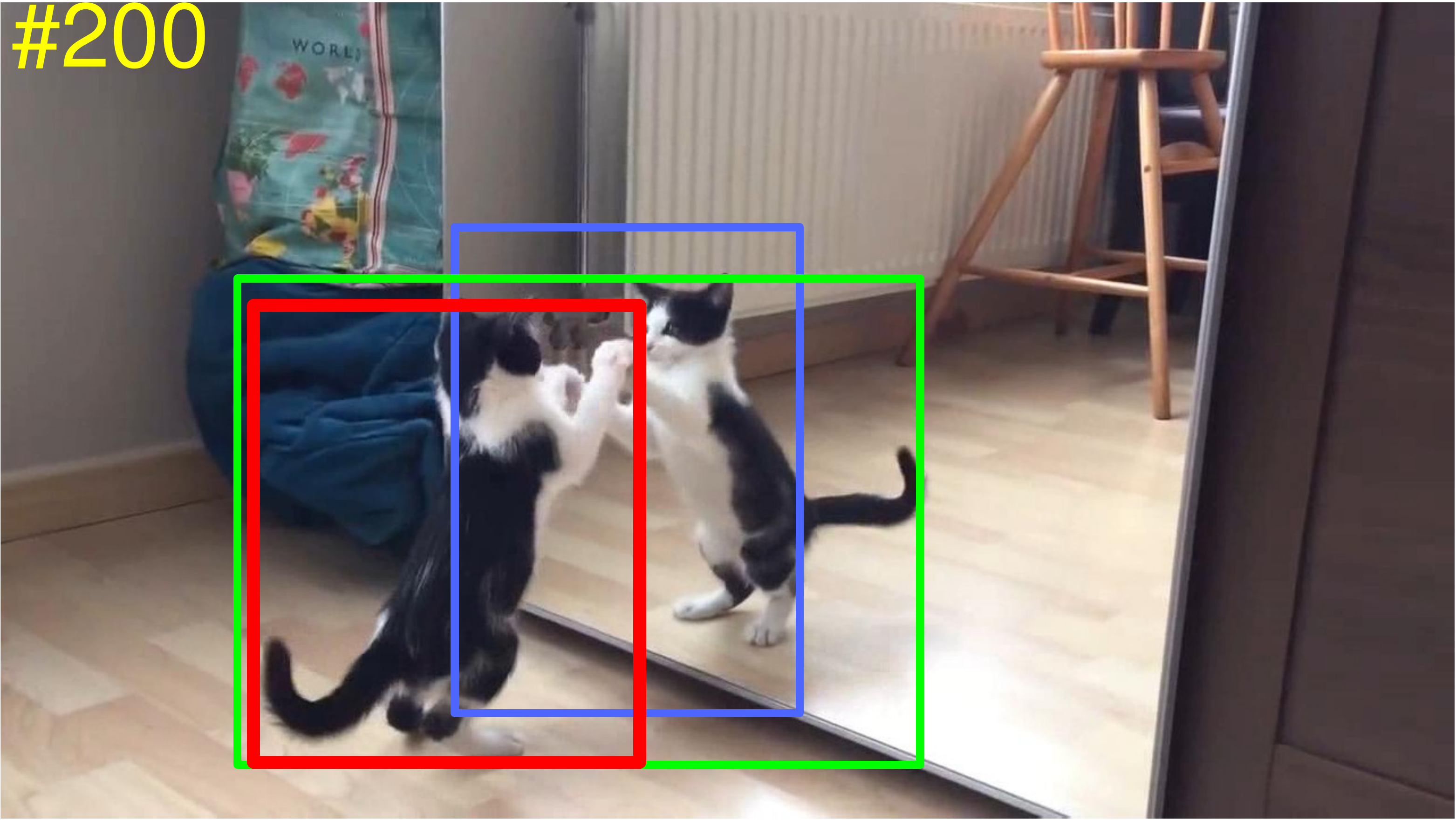}%
	\includegraphics*[trim = 0 0 0 0, width = \wid]{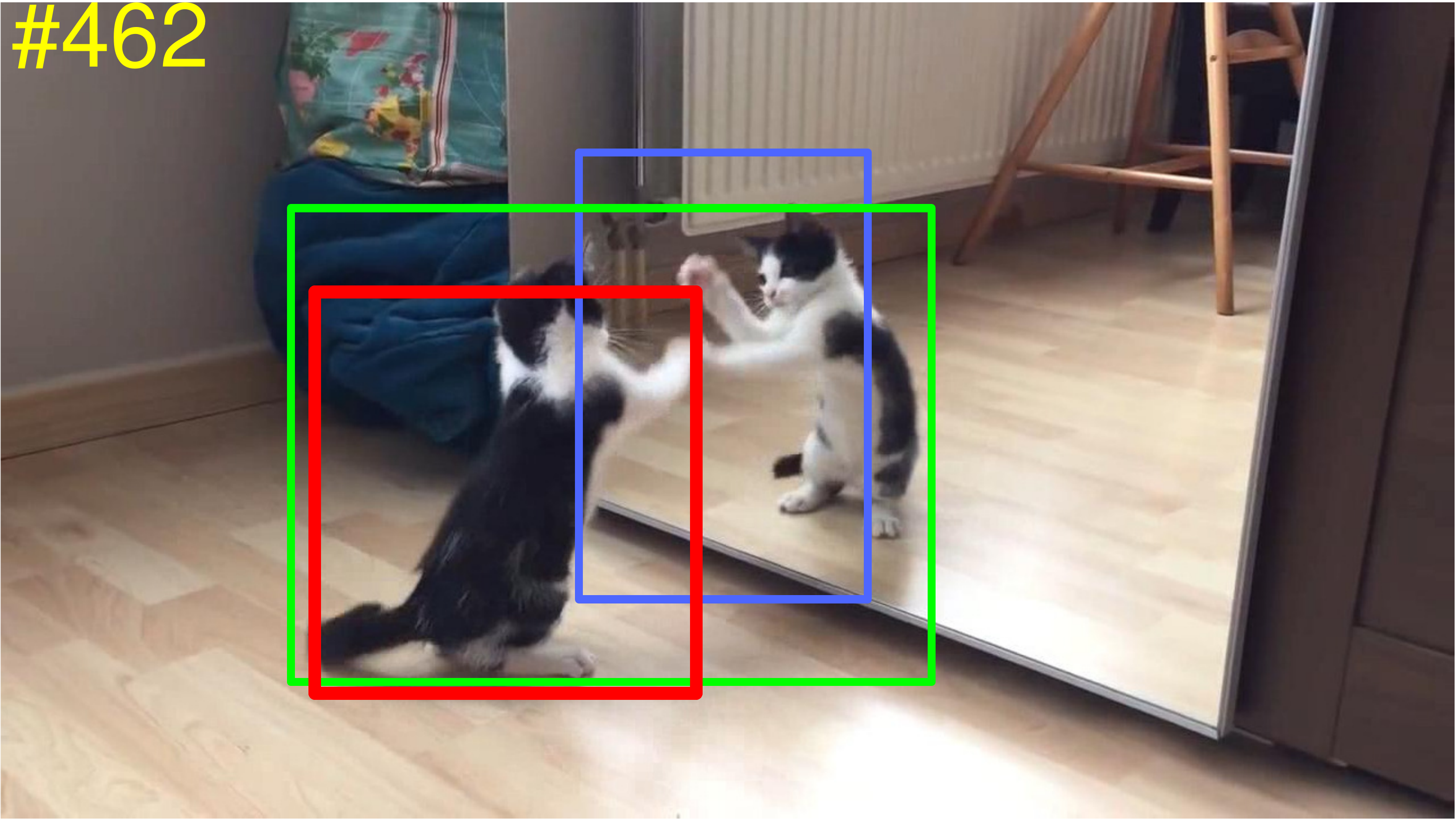}
	\includegraphics*[trim = 0 0 0 0, width = \wid]{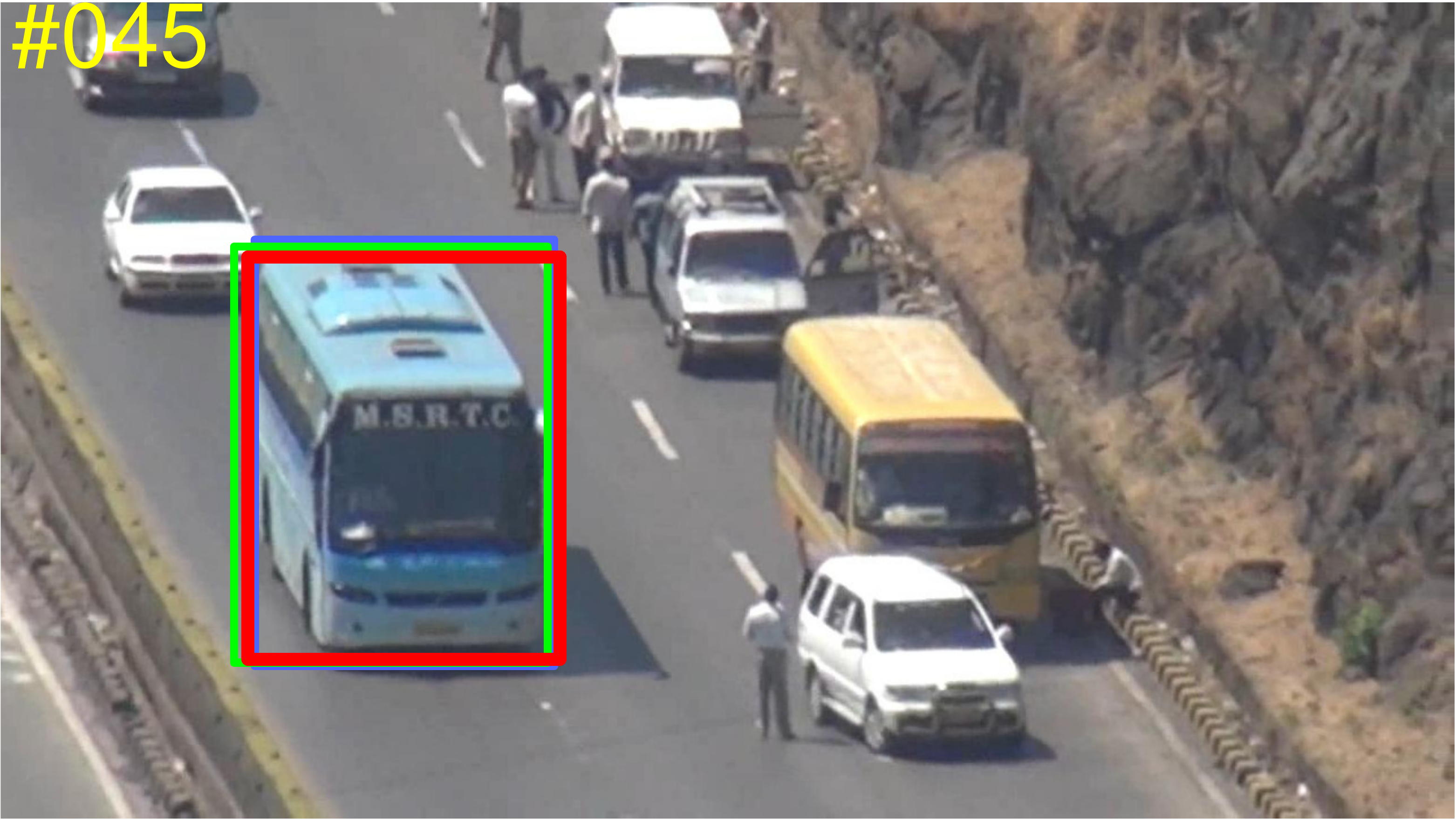}%
	\includegraphics*[trim = 0 0 0 0, width = \wid]{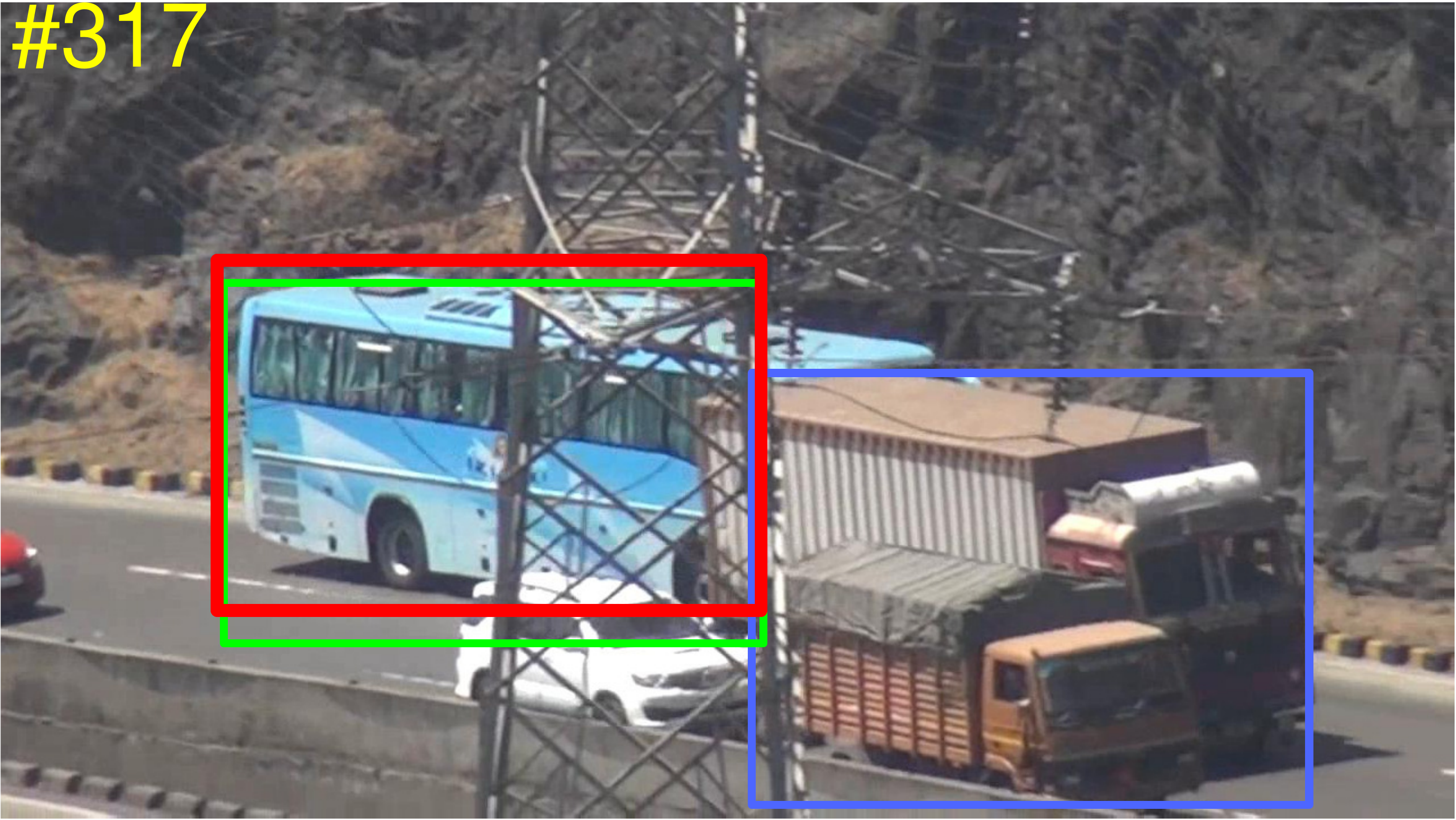}%
	\includegraphics*[trim = 0 0 0 0, width = \wid]{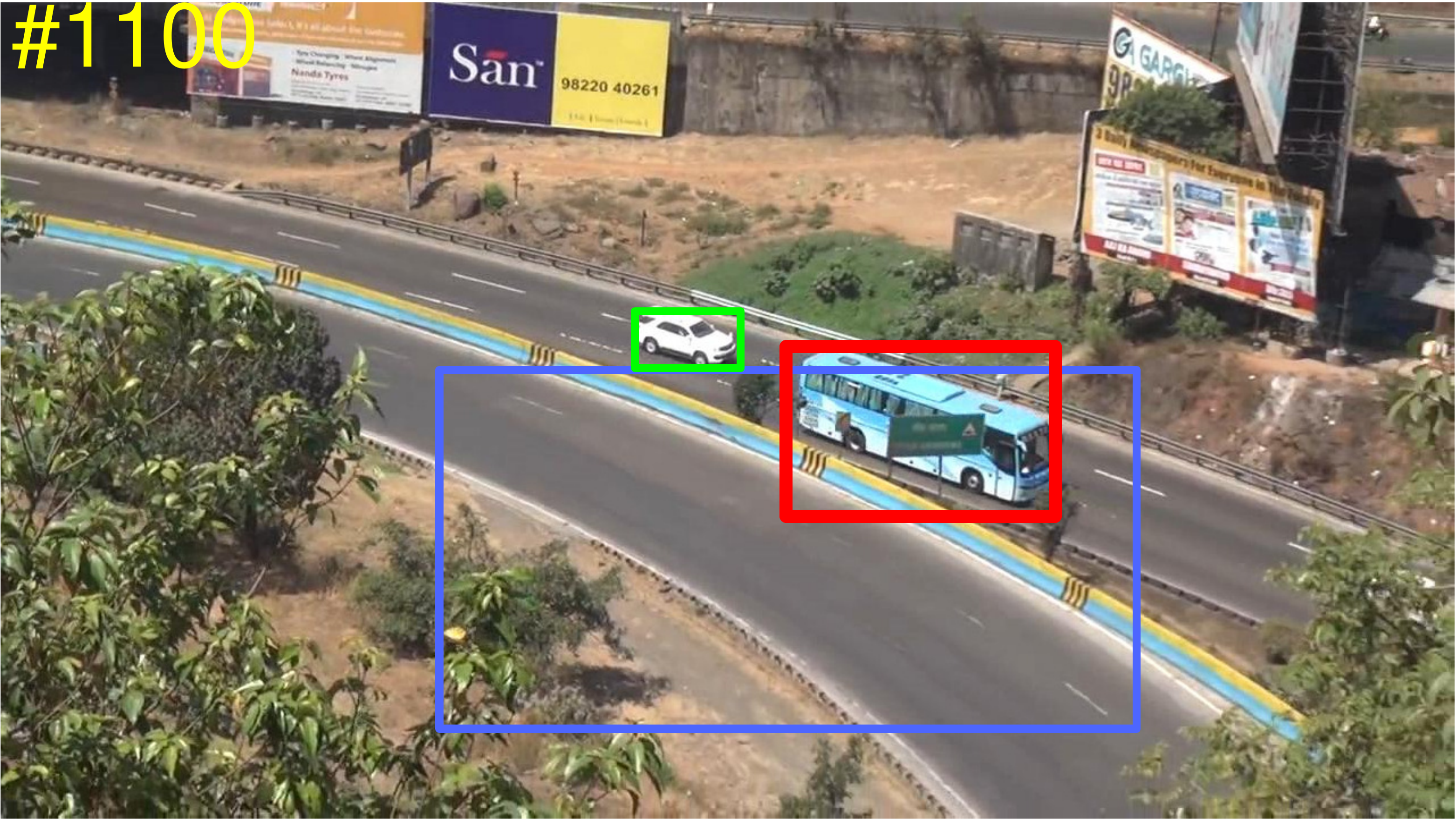}
	\includegraphics*[trim = 0 0 0 0, width = \wid]{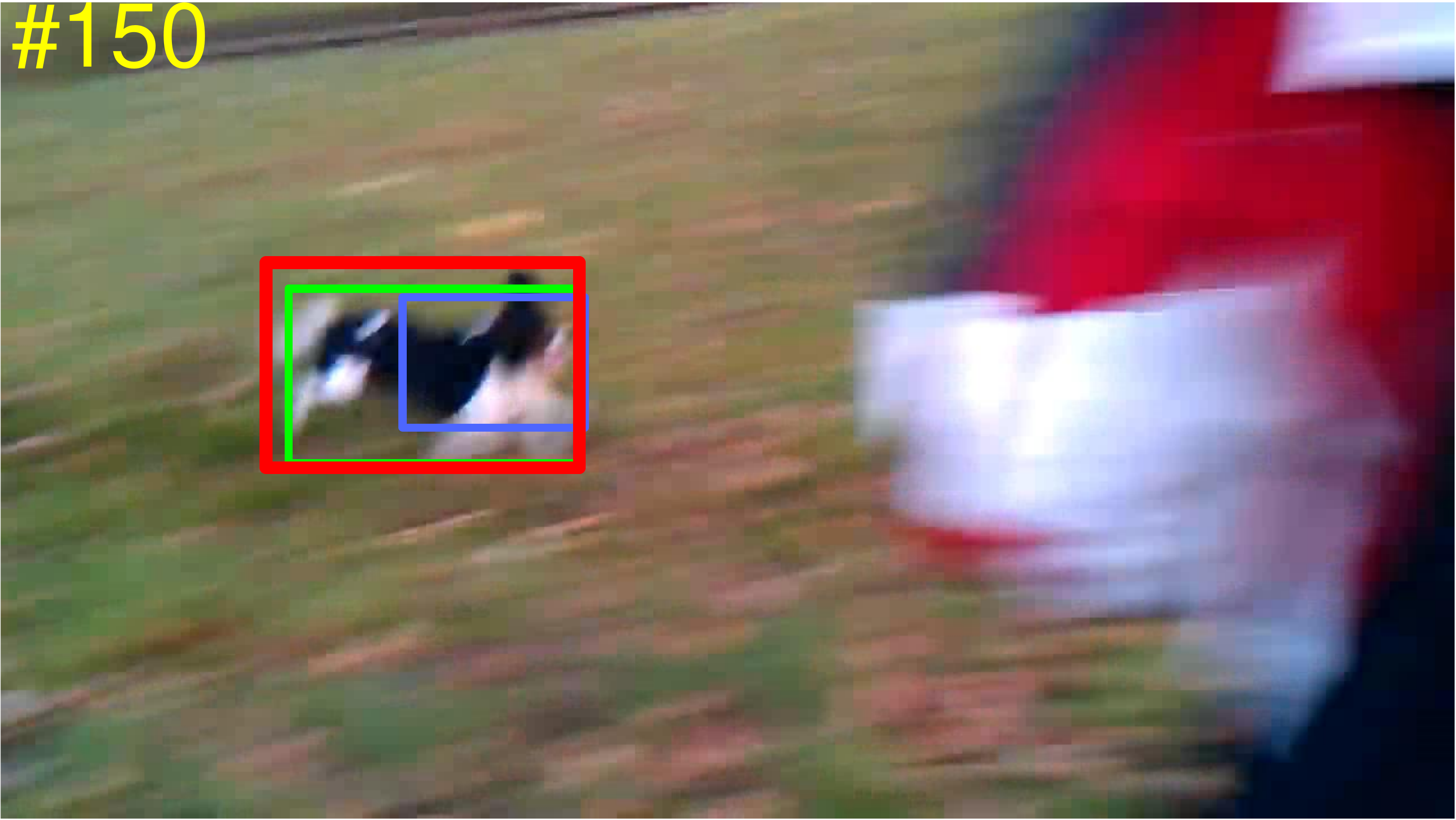}%
	\includegraphics*[trim = 0 0 0 0, width = \wid]{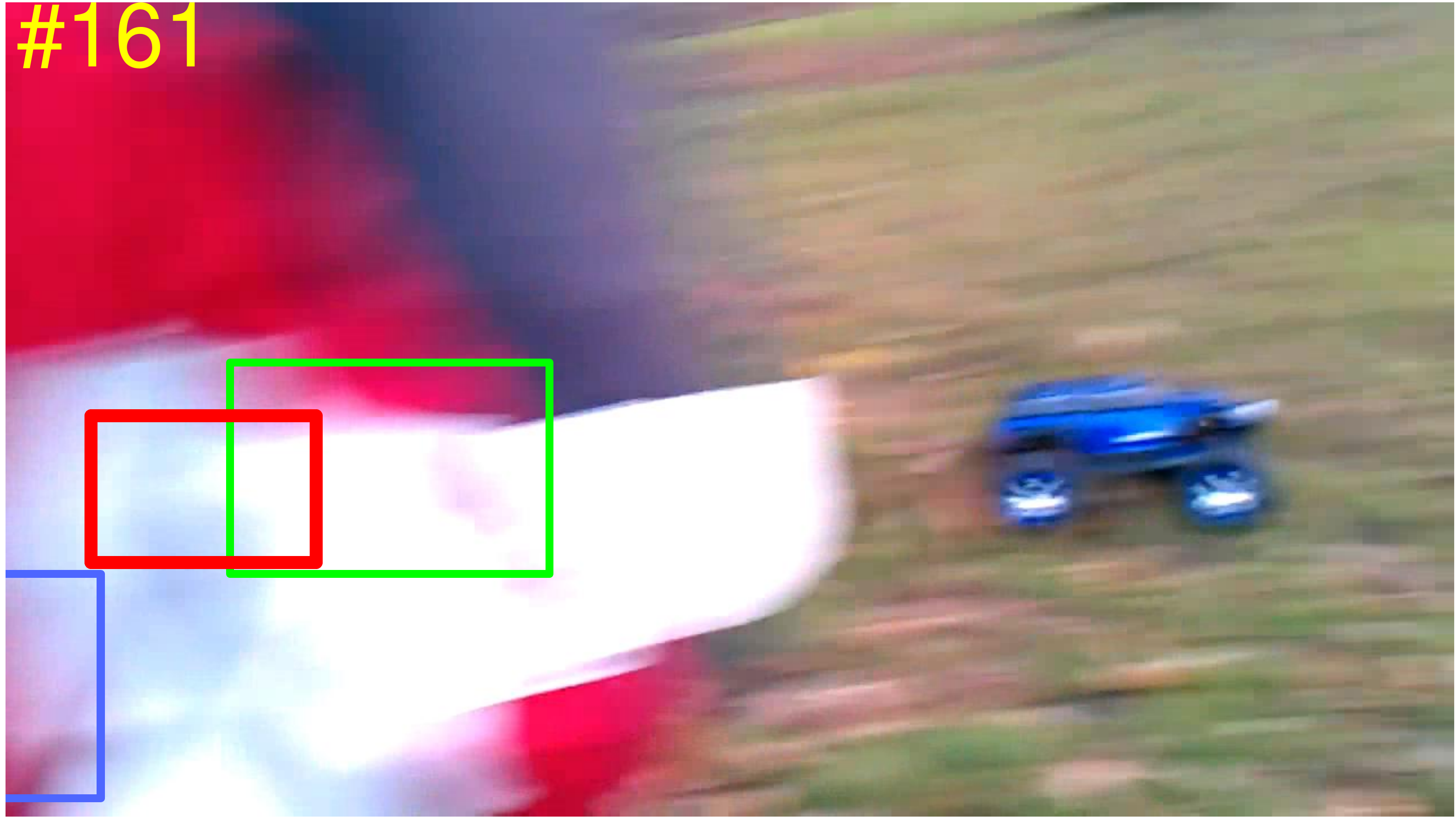}%
	\includegraphics*[trim = 0 0 0 0, width = \wid]{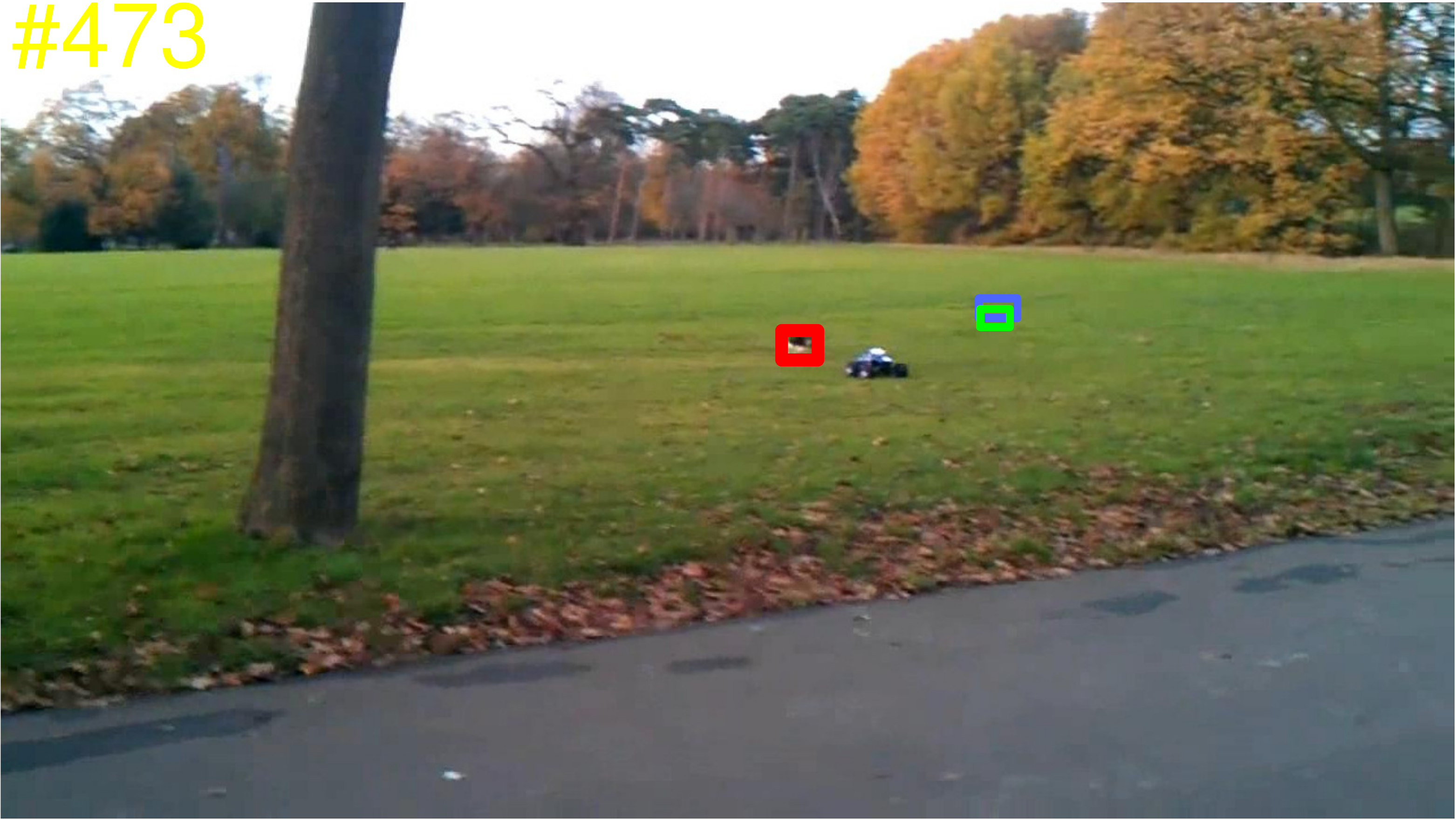}
	\includegraphics*[trim = 0 30 0 0, width = \wid]{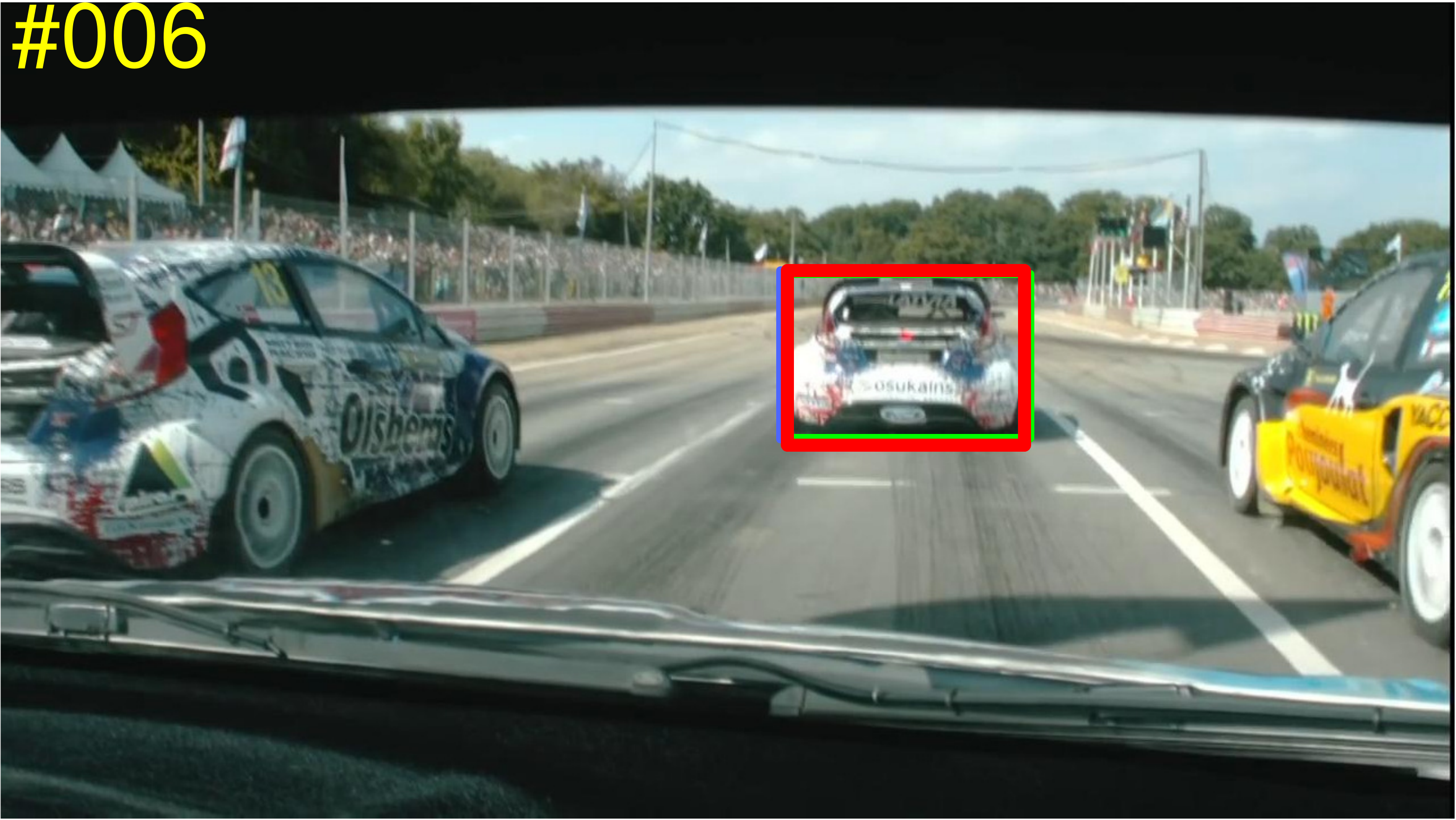}%
	\includegraphics*[trim = 0 30 0 0, width = \wid]{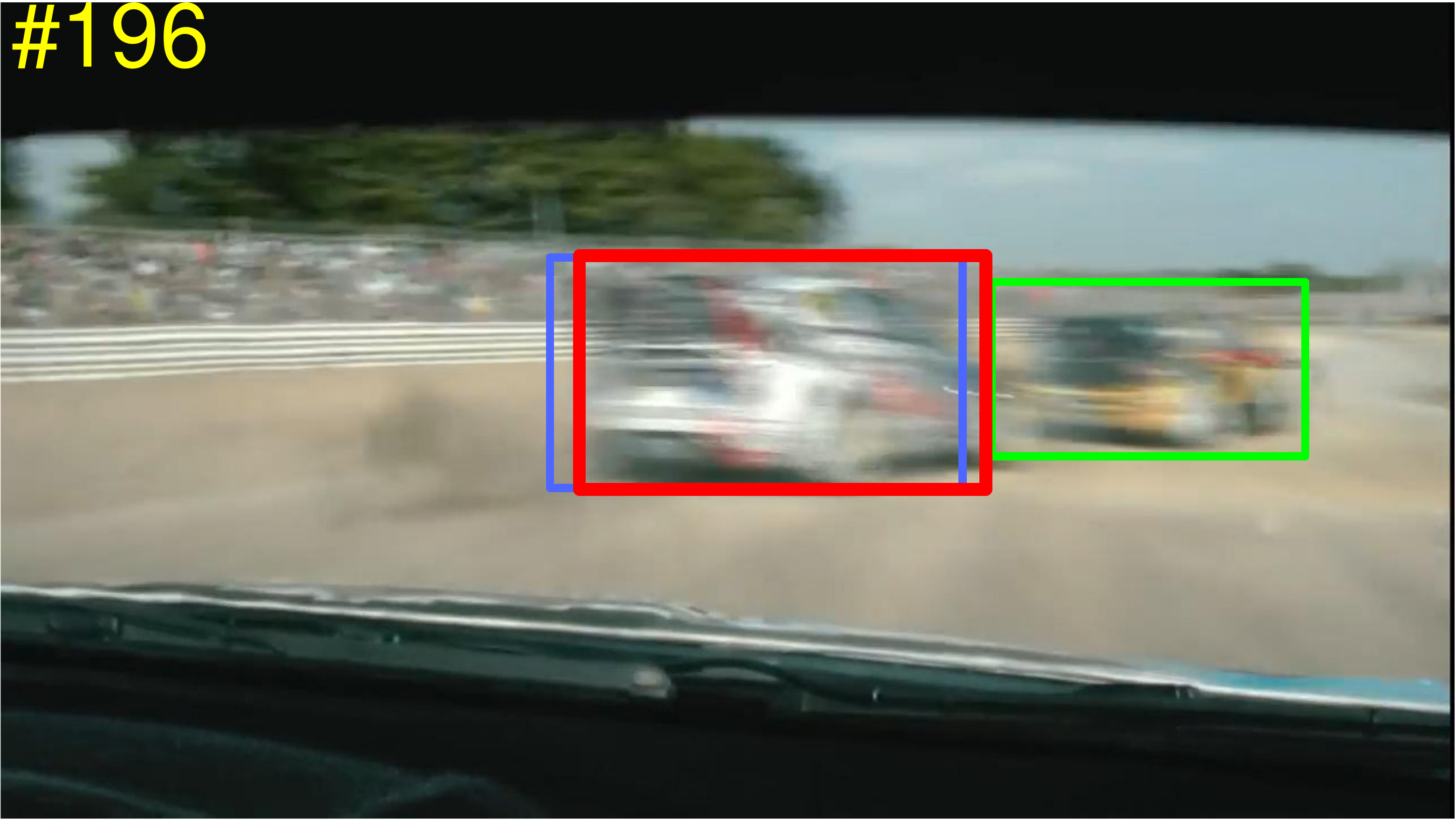}%
	\includegraphics*[trim = 0 30 0 0, width = \wid]{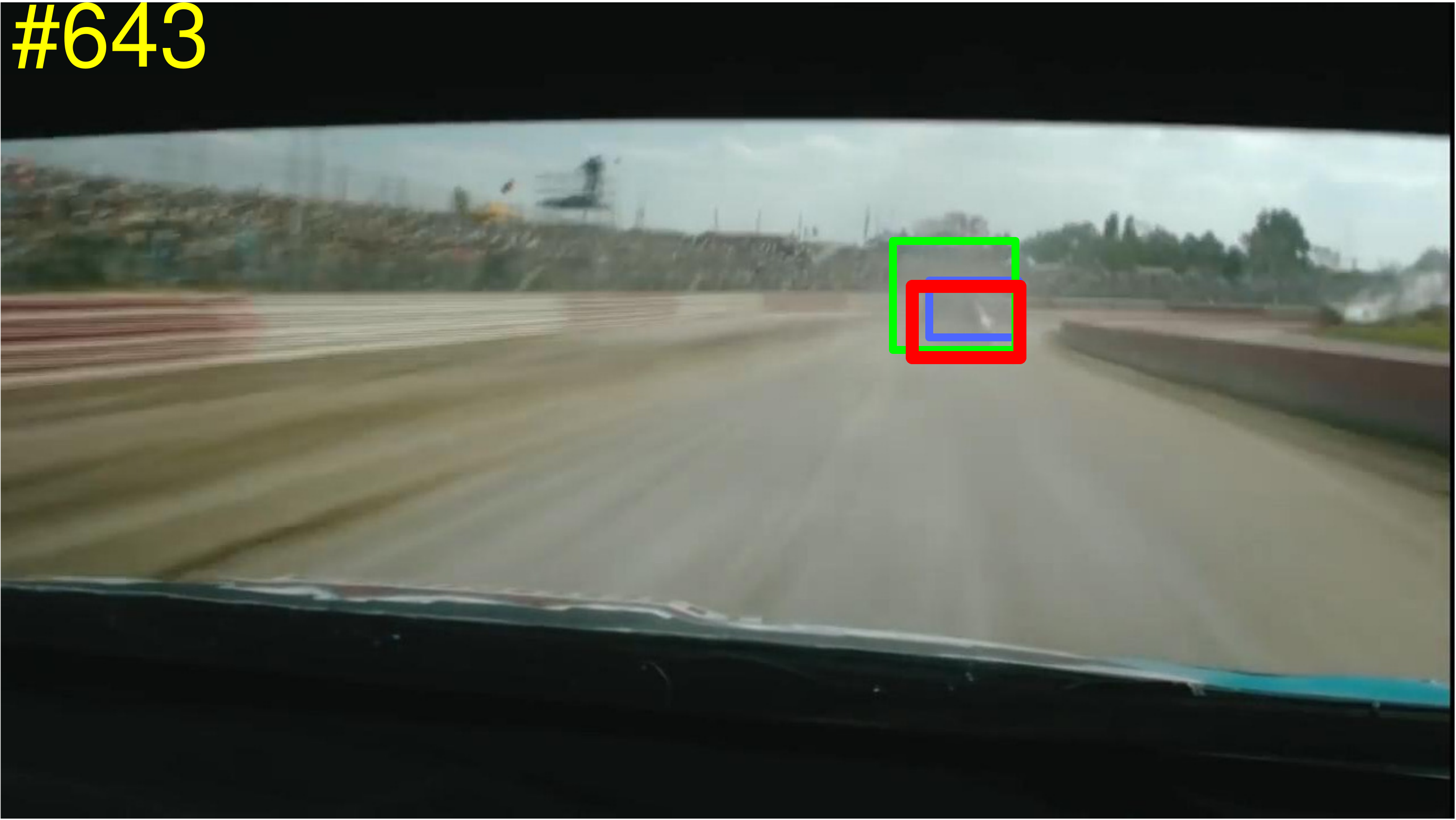}
	\vspace{-0.2mm}
	\includegraphics*[trim = 2 2 2 6, width = 0.85\columnwidth]{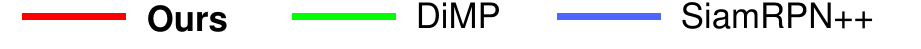}\vspace{-1.5mm}%
	\caption{A comparison of our approach with state-of-the-art trackers DiMP~\cite{DiMP} and SiamRPN++~\cite{SiamRPN++}. In tracking, estimating the uncertainty of the target state is important in the presence of similar objects (first row), occlusions (second row), when determining failures (third row), and in cases of blur or other obstructions (bottom row). Unlike current state-of-the-art, our approach predicts a \emph{probability distribution} $p(y|x)$ of the target state $y$ conditioned on the input image $x$, providing a clear interpretation of the output. The proposed probabilistic formulation further improves the overall performance of the tracker, including the cases shown above.
	}\vspace{-3mm}%
	\label{fig:intro}%
\end{figure}

The aforementioned confidence-based regression strategy is shared by the previously dominant Discriminative Correlation Filter (DCF) paradigm~\cite{MOSSE2010,DanelljanICCV2015,DanelljanECCV2016,GaloogahiCVPR2015,Henriques14,HCF_ICCV15,DRT} and the more recent Siamese trackers~\cite{SiameseFC,Gua2017DSiam,SiamRPN++,SiamRPN,RASNet,UpdateNet}. Both employ a convolutional operation to predict a target confidence $s(y,x)$ at each spatial position $y$, in order to localize the target. Recent work~\cite{DiMP,ATOM} also demonstrated the effectiveness of training a network branch to predict the confidence $s(y,x)$ of the entire target box $y$ to achieve highly accurate bounding box regression. Due to the vast success of these confidence-based regression techniques, we first set out to unify much of the recent progress in visual tracking under this general view.

One distinctive advantage of confidence-based regression is its ability to flexibly represent \emph{uncertainties}, encoded in the predicted confidence values $s(y,x)$. In contrast, a direct regression strategy $y=f(x)$ forces the network to commit to a single prediction $y$, providing no other information. However, the confidence value $s(y,x)$ itself has no clear interpretation, since it simply acts as a quantity to be maximized. The range of values and characteristic of the predicted confidence largely depends on the choice of loss and strategy for generating the corresponding pseudo labels for training. This provides severe challenges when designing strategies for estimating and reasoning with the uncertainty in the prediction. Such measures are highly relevant in tracking, \eg to decide whether to update, if the target is lost, or how uncertain the output is (see Figure~\ref{fig:intro}). We aim to address these limitations by taking a probabilistic view.

\parsection{Contributions}
We propose a formulation for learning to predict the conditional probability density $p(y|x)$ of the target state $y$ given an input image $x$. Unlike the confidence value $s(y,x)$, the density $p(y|x)$ has a clear and direct interpretation, allowing the computation of absolute probabilities. We assume no particular family of distributions, such as Gaussian, instead letting $p(y|x)$ be directly parametrized by the network architecture itself. Specifically, the density $p(y|x)$ is represented by the continuous generalization of the SoftMax operation, previously employed in energy-based models~\cite{LeCun06atutorial} and recently in~\cite{DCTD}. In contrast to these previous works, we also model the uncertainty in the annotations themselves. This is shown to be crucial in visual tracking to counter noise in the annotations and ambiguities in the regression task itself. The network is trained by minimizing the Kullback-Leibler divergence between the predicted density and the label distribution.

We demonstrate the effectiveness of our general approach by integrating it into the recent state-of-the-art tracker DiMP~\cite{DiMP}. Our resulting tracker does not only allow a fully probabilistic representation $p(y|x)$ of the predicted target state. Comprehensive experiments on \emph{seven} benchmark datasets show that our probabilistic representation and training significantly \emph{improves} the performance of the tracker. Our Probabilistic DiMP (PrDiMP) outperforms previous state-of-the-art by a large margin, particularly on available large-scale datasets, including LaSOT ($+2.9\%$ AUC) and TrackingNet ($+1.8\%$ Success).

\section{Regression by Confidence Prediction}
\label{sec:confidence-regression}
In machine learning, regression is fundamentally the problem of learning a mapping $f_\theta: \mathcal{X} \rightarrow \mathcal{Y}$ from an input space $\mathcal{X}$ to a \emph{continuous} output space $\mathcal{Y}$, given a set of example pairs $\{(x_i,y_i)\}_i \subset \mathcal{X} \times \mathcal{Y}$. For our purposes, $\mathcal{X}$ constitutes the space of images. The most straight-forward take on regression is to directly learn the function $f_\theta$, parametrized as \eg a deep neural network with weights $\theta$, by minimizing a loss $L(\theta) = \sum_i \ell(f_\theta(x_i), y_i)$. Here, the function $\ell$ measures the discrepancy between the prediction $f_\theta(x_i)$ and corresponding ground-truth value $y_i$. While the choice of loss $\ell$ is highly problem dependent, popular alternatives include the $L^p$ family, $\ell(y, y') = \|y - y'\|_p^p$.

\subsection{General Formulation}
While direct regression is successfully applied for many computer vision problems, including optical flow~\cite{pwcnet} and depth estimation~\cite{monodepth}, it has proven less suitable for other vision tasks. Examples of the latter include visual tracking~\cite{SiameseFC,danelljan2016discriminative,Henriques14}, object detection~\cite{IOUNet,law2018cornernet,Ren2015FasterRT} and human pose estimation~\cite{cao2017realtime,sun2019deep,xiao2018simple}. In these problems, networks are often trained to instead predict a confidence score, which is then maximized in order to achieve the final estimate. Confidence prediction has prevailed over standard direct regression in these circumstances thanks to two key advantages. First, confidence prediction can capture the presence of uncertainties, multiple hypotheses and ambiguities in the output space $\mathcal{Y}$. The network does not have to commit to a single estimate $f_\theta(x) = y$. Second, the network can more easily exploit symmetries shared by $\mathcal{X}$ and $\mathcal{Y}$, such as translational invariance in the case of image 2D-coordinate regression tasks, which are particularly suitable for CNNs.

Formally, we define confidence-based regression as learning a function $s_\theta: \mathcal{Y}\times\mathcal{X} \rightarrow \reals$ that predicts a scalar confidence score $s_\theta(y,x) \in \reals$ given an output-input pair $(y,x)$. The final estimate $f(x) = y^*$ is obtained by maximizing the confidence \wrt to $y$,
\begin{equation}
\label{eq:confidence-regression}
f(x) = \argmax_{y\in \mathcal{Y}} s_\theta(y,x) \,.
\end{equation}
The regression problem is thus transformed to learning the function $s_\theta$ from the data $\{(x_i,y_i)\}_i$. This is generally performed by defining a function $a: \mathcal{Y} \times \mathcal{Y} \rightarrow \reals$ for generating a pseudo label $a(y,y_i)$, acting as the ground-truth confidence value for the prediction $s_\theta(y,x_i)$. The confidence prediction network can then be trained by minimizing the loss $L = \sum_i L(\theta; x_i, y_i)$, where
\begin{equation}
\label{eq:confidence-loss}
L(\theta; x_i, y_i) = \int_\mathcal{Y} \ell\big(s_\theta(y,x_i), a(y,y_i)\big) \diff y \,.
\end{equation}
The function $\ell : \reals \times \reals \rightarrow \reals$ now instead measures the discrepancy between the predicted confidence value $s_\theta(y,x_i)$ and the corresponding label value $a(y,y_i)$. In practice, a variety of losses $\ell$ and pseudo label functions  $a$ are employed, depending on the task at hand. In the next section, some more popular examples are studied, where our discussion focuses on the visual tracking problem in particular.

\subsection{In Visual Tracking}
\label{sec:confidence-tracking}
In visual tracking, the task is to regress the state of the target object in each frame of the video, given its initial location. The state is most often represented as an axis-aligned bounding box $y \in \reals^4$. Compared to other vision tasks, this problem is particularly challenging since an example appearance of the target object is only provided at test-time. The tracker must thus learn a model based on this first example in order to locate the object in each frame.

Due to this challenging nature of the problem, the majority of approaches until very recently, focused on regressing the center 2D image coordinate $y \in \reals^2$ of the target object, and then optionally using this model to estimate the one-parameter scale factor by a multi-scale search. This class of methods include the widely popular Discriminative Correlation Filter (DCF) approaches~\cite{MOSSE2010,DanelljanICCV2015,DanelljanECCV2016,Henriques14}, most of the more recent Siamese networks~\cite{SiameseFC,Gua2017DSiam,RASNet,UpdateNet} and other earlier approaches~\cite{MEEM2014}. The formulation \eqref{eq:confidence-regression}, \eqref{eq:confidence-loss} is explicitly utilized in the theory of Structural SVMs~\cite{SSVM}, employed in the well-known Struck tracker~\cite{struck}. In DCF-based methods, a convolutional layer is trained \emph{online}, \ie during tracking, to predict a target confidence score
\begin{equation}
\label{eq:score-conv}
s_\theta(y,x) = (w_\theta \conv \phi(x))(y) \,.
\end{equation}
Here, $w_\theta$ is the convolution kernel and $\phi(x)$ are the features extracted from the image $x$, typically by a CNN with frozen weights~\cite{DanelljanECCV2016,HCF_ICCV15}. The result of the convolution \eqref{eq:score-conv} is evaluated at the spatial location $y$ to obtain the confidence $s_\theta(y,x)$. The DCF paradigm adopts a squared loss $\ell(s,a) = (s - a)^2$ on the confidence predictions, which enables efficient optimization of \eqref{eq:confidence-loss} \wrt $w_\theta$ in the Fourier domain~\cite{MOSSE2010,DanelljanECCV2016}. Nearly all DCF methods employ a Gaussian confidence pseudo label $a(y,y_i) = e^{-\frac{\|y - y_i\|^2}{2\sigma^2}}$ centered at the target position $y_i$ in frame $x_i$.

In contrast to DCF, Siamese trackers~\cite{SiameseFC,Gua2017DSiam,SiamRPN++,SiamRPN,RASNet,UpdateNet} aim to fully learn the parameters $\theta$ of the network in an offline training stage. This is performed by learning an embedding space $\phi_\theta$ in which similarities between a target template $z$ and frame $x$ can be computed as a correlation,
\begin{equation}
\label{eq:siam-corr}
s_\theta(y,x) = (\phi_\theta(z) \conv \phi_\theta(x))(y) \,.
\end{equation}
Siamese methods often employ a binary cross entropy loss 
\begin{equation}
\label{eq:bce}
\ell(s,a) = a\log(1 + e^{-s}) + (1-a)\log(1 + e^{s})
\end{equation}
in \eqref{eq:confidence-loss} to train the network parameters $\theta$. That is, target localization is treated as a dense binary classification problem, where the pseudo label $a(y,y_i) \in [0,1]$ represents the target/background class, or more generally, a Bernoulli distribution. It is commonly set to $a(y,y_i) = 1$ in the target vicinity $\|y - y_i\| < r$ and $a(y,y_i) = 0$ otherwise~\cite{SiameseFC}.

To achieve an accurate prediction of the full target bounding box, a few recent trackers~\cite{DiMP,ATOM,SiamRPN++,SiamRPN} have achieved remarkable performance by separating the tracking problem into two parts. First, the object is coarsely localized using techniques reminiscent of the aforementioned approaches, that are robust to similar background objects, clutter and occlusions. In the second stage, a separate network branch is employed for regressing the target bounding box. 
For this purpose, the ATOM tracker~\cite{ATOM} employs an IoU-Net~\cite{IOUNet} based network head $s_\theta(y,x)$, which scores any input bounding-box $y\in \reals^4$. This head is trained in an offline learning stage to predict the Intersection-over-Union (IoU) overlap $a(y,y_i) = \text{IoU}(y,y_i)$ using the squared error $\ell(s,a) = (s - a)^2$ in \eqref{eq:confidence-loss}. In this case, the integral \eqref{eq:confidence-loss} is approximated by sampling bounding boxes during training. During tracking, the optimal box \eqref{eq:confidence-regression} is achieved by gradient-based maximization of the predicted confidence. 

More recently, Bhat~\etal~\cite{DiMP} proposed the DiMP tracker by designing a meta-learning based network architecture which predicts discriminative target model weights $w_\theta = \psi_\theta(\{(\phi_\theta(z_j), y_j)\}_j)$ in \eqref{eq:score-conv} from a set of sample pairs $\{(z_j, y_j)\}_j$. The predicted weights are then employed for the first-stage robust target localization, and updated during tracking through a learned recurrent optimization process. The target model predictor $\psi_\theta$ is learned end-to-end using a robust version of the squared error $\ell$ and Gaussian confidence labels $a(y,y_i)$. For the second stage, it adopts the bounding-box regression technique proposed in ATOM.

\section{Method}
\label{sec:approach}

We propose a probabilistic regression model that integrates all advantages of confidence-based regression. However, unlike the aforementioned confidence-based models, our approach generates a predictive probability distribution $p(y|x_i, \theta)$ as output. The network is trained by minimizing the KL divergence between the predictive density $p(y|x, \theta)$ and the conditional ground-truth distribution $p(y|y_i)$, which models label noise and ambiguities in the task itself. During inference, a point estimate of the regressed value is obtained by maximizing the predicted density.

Our approach possesses a few important advantages compared to the confidence-based regression methods.
In the latter, the prediction $s_\theta(y,x)$ can be difficult to interpret, and its value largely depend on the pseudo label function $a$ and employed loss $\ell$. In contrast, the probabilistic nature of our method allows reasoning about the uncertainty in the output. Moreover, in our approach, the pseudo label function $a$ is replaced by the label-conditional distribution $p(y|y_i)$, which models noise and uncertainty in the annotation $y_i$. Lastly, in contrast to confidence-based regression, our approach does not require a choice of loss $\ell$. Instead, we directly minimize the Kullback-Leibler (KL) divergence between the predictive distribution and the ground-truth. Next, we provide a general formulation of the proposed regression model, and apply it to tracking in Section~\ref{sec:tracking}.

\begin{figure}[!t]
	\centering%
	\includegraphics*[trim = 0 0 0 50, width = \columnwidth]{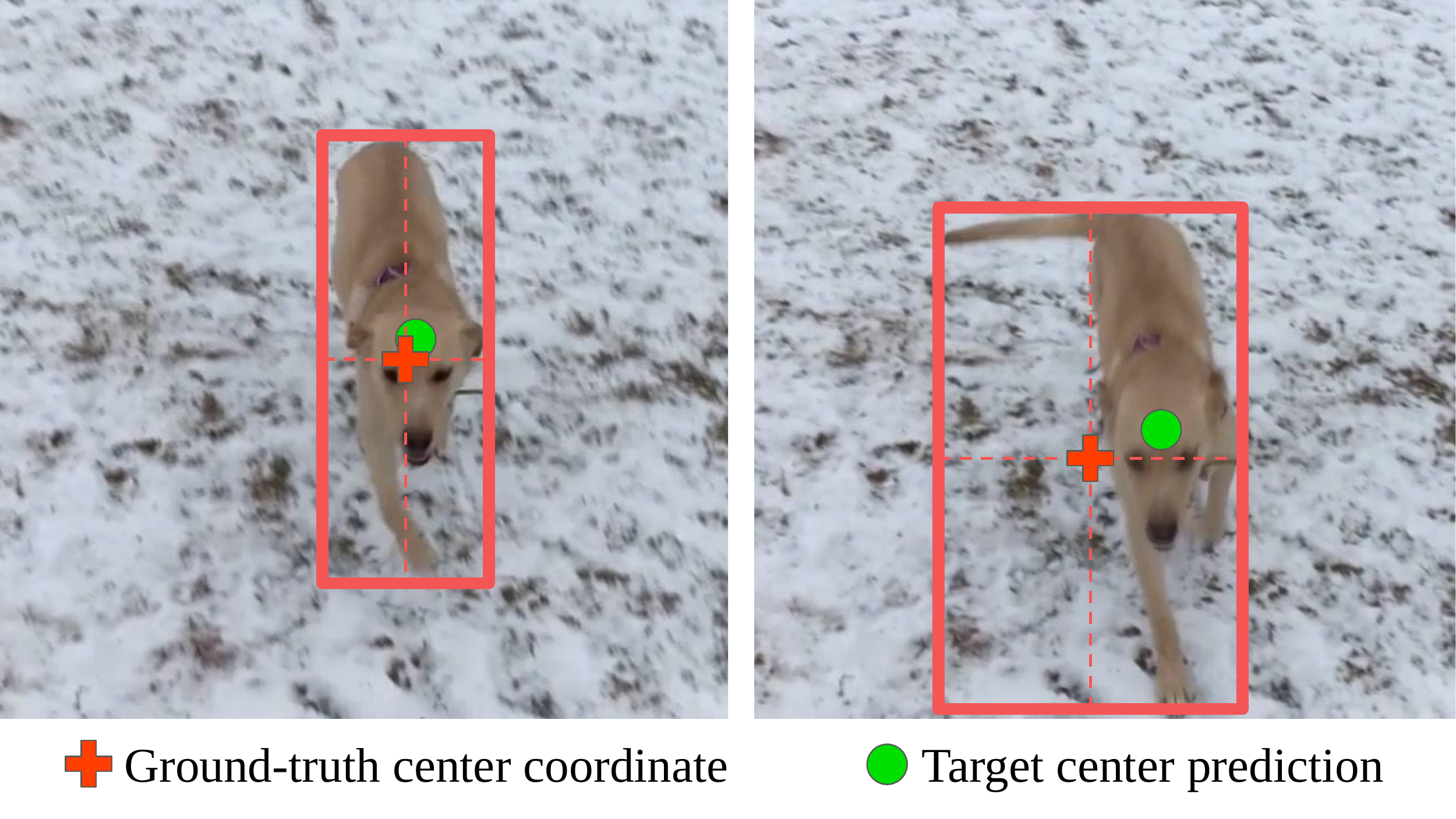}%
	\vspace{-2mm}%
	\caption{Trackers are most often trained to predict the center coordinate of the ground-truth bounding box (red). This is a natural choice for the left frame and aligns well with the tracker prediction (green). Only two frames later (right), the motion of the tail has led to a radical shift in the ground-truth center location, which now lies in the background. This is not necessarily a natural definition of the target center coordinate, due to the minor change in the object appearance. Target center regression is thus an ambiguous task, where it is unclear how to define the ``correct'' value $y_i$. Our formulation models such ambiguity and uncertainty in the regression task by a distribution $p(y|y_i)$ of ``correct'' values.
	}\vspace{-3mm}%
	\label{fig:ambiguity}%
\end{figure}

\subsection{Representation}
In this section, we formulate an approach for effectively training the network to predict a probability distribution $p(y|x, \theta)$ of the output $y$ given the input $x$. The density itself is represented using the formulation previously employed in probabilistic energy-based deep learning~\cite{LeCun06atutorial} and the recent deep conditional target densities~\cite{DCTD},
\begin{equation}
\label{eq:pred}
p(y|x, \theta) = \frac{1}{Z_\theta(x)} e^{s_\theta(y,x)} \,,\; Z_\theta(x) = \int e^{s_\theta(y,x)} \diff y \,.
\end{equation}
As for the confidence-based methods described in Section~\ref{sec:confidence-regression}, $s_\theta: \mathcal{Y}\times\mathcal{X} \rightarrow \reals$ is a deep neural network mapping the output-input pair $(y,x)$ to a scalar value. The expression \eqref{eq:pred} converts this value to a probability density by exponentiation and division by the normalizing constant $Z_\theta(x)$. In fact, it should be noted that \eqref{eq:pred} is a direct generalization of the SoftMax operation to an arbitrary output space $\mathcal{Y}$. 

Since the output of the network represents a probability density over $\mathcal{Y}$, we can learn the network parameters $\theta$ by applying techniques for fitting a probability distribution to data. Given training sample pairs $\{(x_i, y_i)\}_i$, the simplest approach is to minimize the negative log-likelihood,
\begin{equation}
\label{eq:nll}
-\log p(y_i|x_i, \theta) = \log\left(\int e^{s_\theta(y,x_i)} \diff y\right) - s_\theta(y_i,x_i) \,.
\end{equation}
This strategy was recently successfully applied for a number of computer vision tasks~\cite{DCTD}, including bounding-box regression in visual tracking.
One advantage of the negative log-likelihood loss \eqref{eq:nll} is that it only employs the training sample $(x_i,y_i)$ itself, without further assumptions. However, this brings an important limitation, discussed next.

\begin{figure}[!b]
	\centering%
	\newcommand{\wid}{22mm}%
	\vspace{-5mm}\includegraphics*[trim = 0 0 20 0, height = \wid]{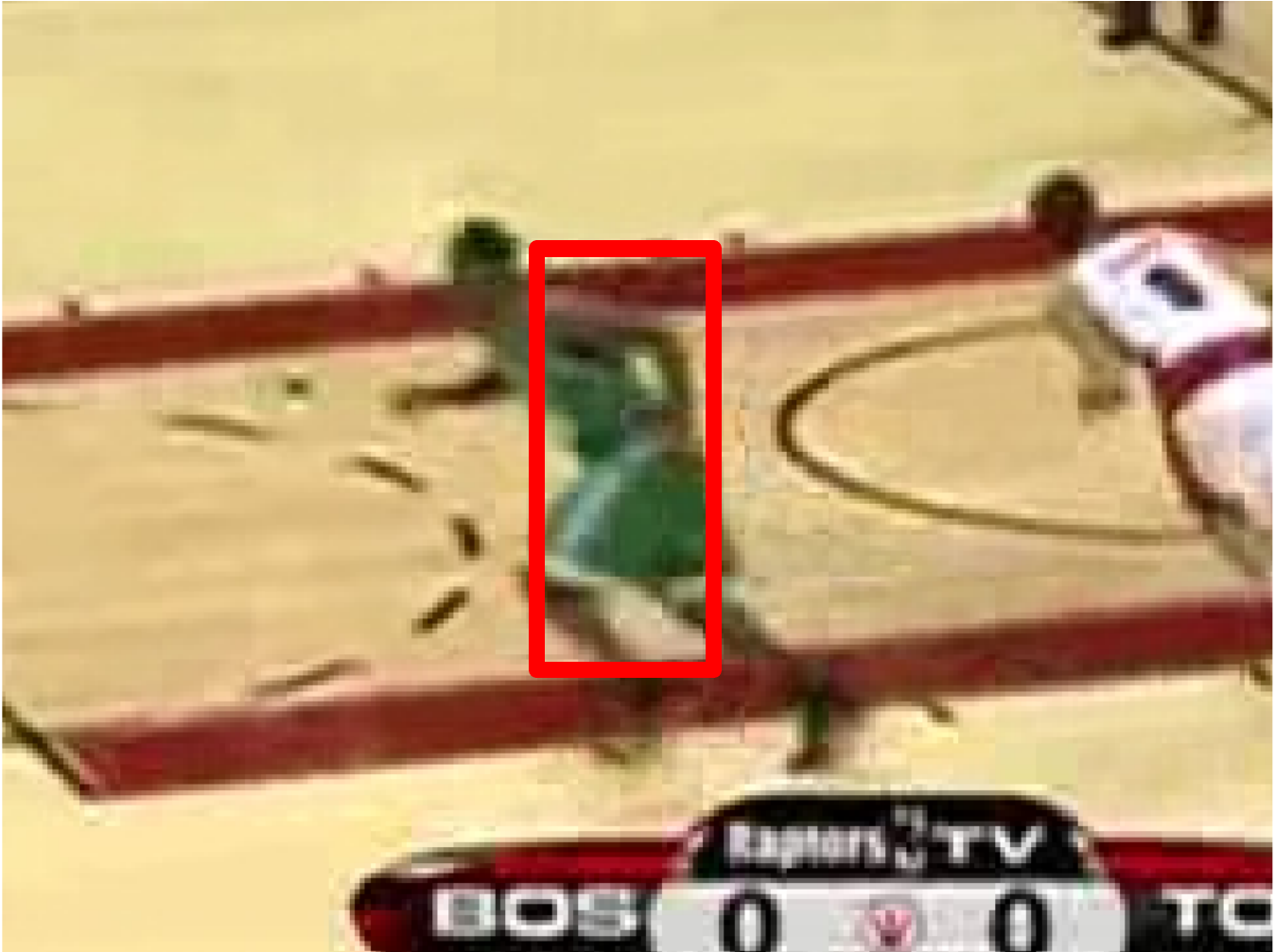}~%
	\includegraphics*[trim = 190 20 190 70, height = \wid]{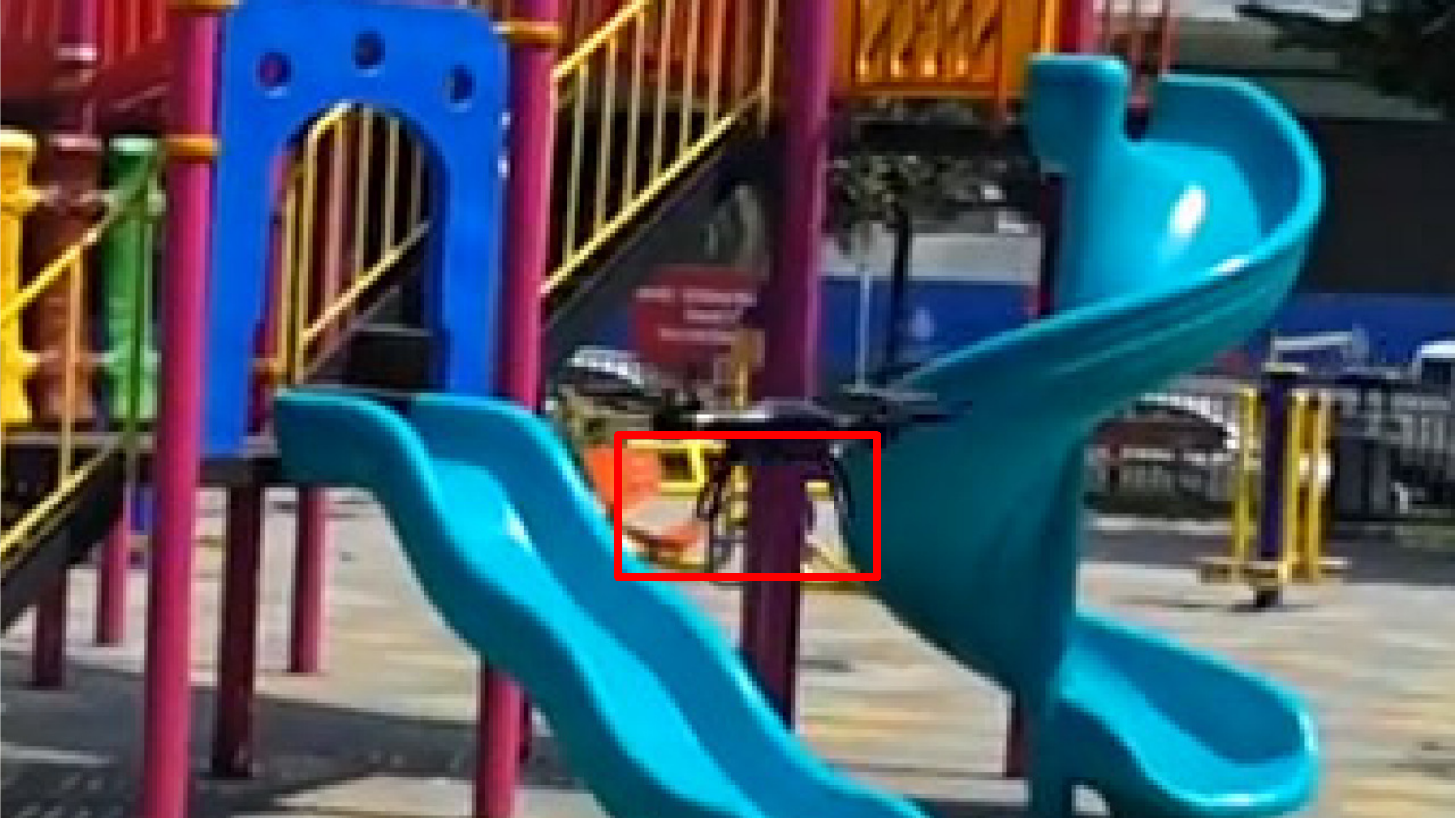}~%
	\includegraphics*[trim = 120 0 160 0, height = \wid]{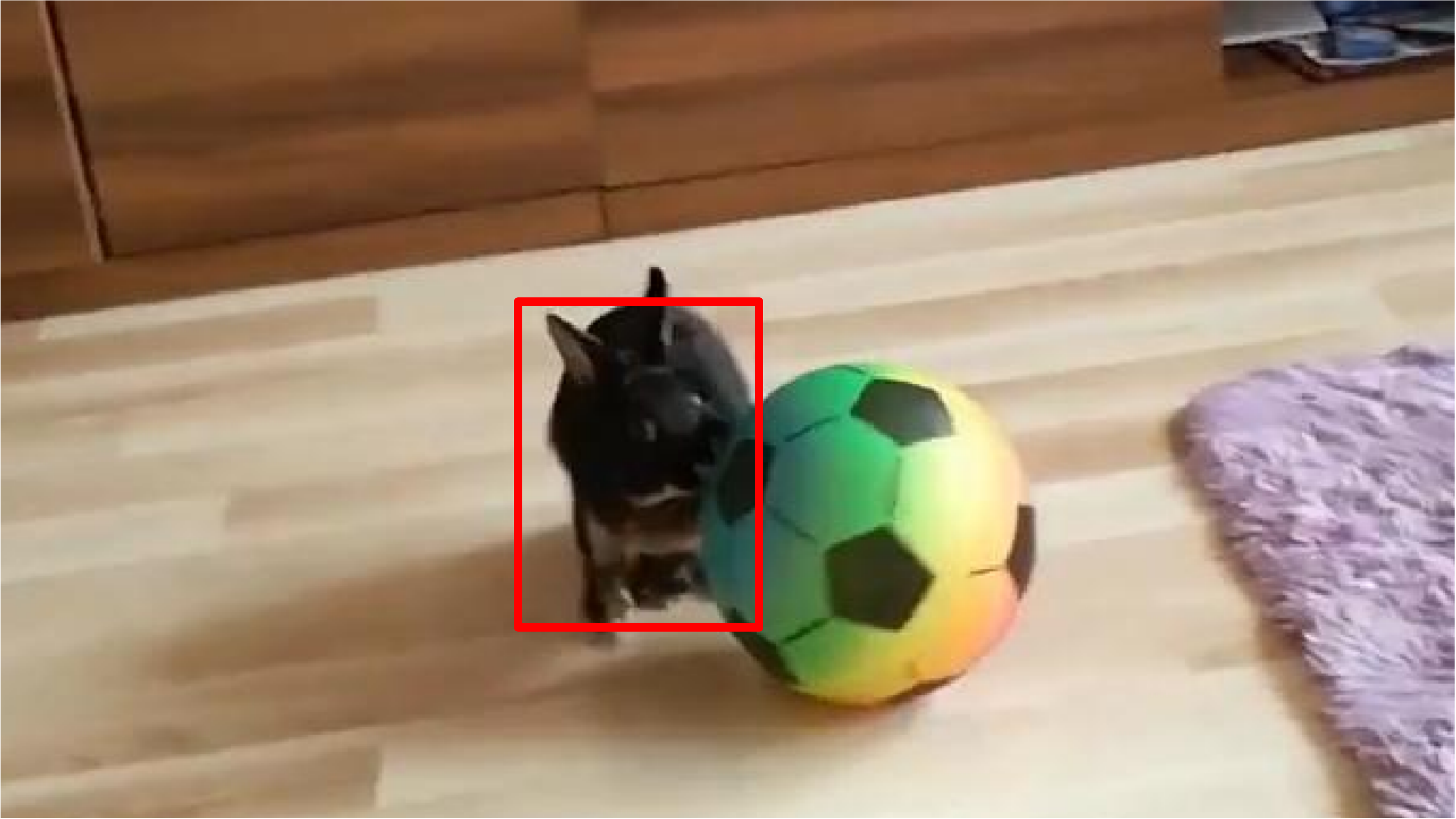}\vspace{-0.5mm}%
	\caption{Examples of noisy, incorrect or ambiguous ground truth bounding box annotations $y_i$ from different datasets. These aspects are modeled by our label distribution $p(y|y_i)$.}
	\label{fig:noisy-anno}%
\end{figure}

\subsection{Label Uncertainty and Learning Objective}
\label{sec:kl-objective}

Compared to the negative log-likelihood loss \eqref{eq:nll}, the confidence-based paradigm described in Section~\ref{sec:confidence-regression} enjoys a certain flexibility stemming from the pseudo label function $a(y,y_i)$. In practice, the design of $a(y,y_i)$ has been shown to be critical for tracking performance~\cite{BhatECCV2018,promoting-dcf-ECCV2016}. We believe that this is mostly due to the inherent ambiguity of the task and the uncertainty in the label $y_i$ itself.
Most tracking approaches focus on regressing the center coordinate $y \in \reals^2$ of the target in the image. However, for most objects, this is an inherently ambiguous and ill-defined task. While the center coordinate can be defined as the center of mass of the target bounding box, this is hardly a visually intuitive definition for a human, or a tracking algorithm for that matter.

Consider the example in Figure~\ref{fig:ambiguity}. When the target dog in the video raises its tail, the center of mass changes radically and ends up at a background pixel. On the other hand, the appearance and location of the object is almost unchanged. The tracker would thus naturally predict a similar target center location as before. This demonstrates that the definition of the target center is largely ambiguous and that the center of mass is often confusing for the tracker. The pseudo label function $a(y,y_i)$ can encapsulate this ambiguity by having a wider high-confidence peak, which has been shown beneficial for training tracking models~\cite{BhatECCV2018}.
Another source of uncertainty is label noise. Accurate bounding box annotation is a difficult task, especially in the presence of occlusions, motion blur, and for small objects, as shown in Figure~\ref{fig:noisy-anno}. In other words, multiple annotators would naturally disagree for a given object, with some level of variation. This variation, or noise, in the annotation is most often ignored when training the network.

We propose to probabilistically model label noise and task ambiguities for the regression problem as a conditional ground-truth distribution $p(y|y_i)$. It characterizes the probability density of the ground-truth output value $y$, given the annotation $y_i$. Instead of the negative log-likelihood \eqref{eq:nll}, we train the network to minimize the KL divergence to $p(y|y_i)$,
\begin{align}
\label{eq:kl-loss}
&\KL(p(\cdot|y_i),p(\cdot|x_i,\theta)) = \int p(y|y_i) \log \frac{p(y|y_i)}{p(y|x_i,\theta)} \diff y \nonumber \\
&\sim \log \left(\int e^{s_\theta(y,x_i)} \diff y\right) - \int s_\theta(y,x_i) p(y|y_i)  \diff y \,.
\end{align}
Here, $\sim$ denotes equality up to a constant term. The second line in \eqref{eq:kl-loss} corresponds to the cross entropy between the two distributions and the discarded constant term is the negative entropy $\int p(y|y_i) \log p(y|y_i) \diff y$ of the label distribution. See Appendix~\ref{sec:klloss} for a detailed derivation.

The loss \eqref{eq:kl-loss} naturally integrates information about the uncertainty $p(y|y_i)$ in the annotated sample $(x_i,y_i)$. Unlike the pseudo label function $a(y|y_i)$ employed in confidence-based regression, $p(y|y_i)$ has a clear interpretation as a probability distribution. In fact, $p(y|y_i)$ could be empirically estimated by obtaining multiple annotations for a small subset of the data. In the case of a Gaussian model $p(y|y_i) = \norm(y|y_i,\sigma^2)$, the variance $\sigma^2$ can be estimated as the empirical variance of these annotations. In this work, we simply consider $\sigma^2$ a hyper-parameter.

\subsection{Training}
\label{sec:training}

In this section, we consider strategies for training the network parameters $\theta$ based on the loss \eqref{eq:kl-loss}. In practice, this requires approximating the two integrals in \eqref{eq:kl-loss}. We consider two techniques for this purpose, namely grid sampling and Monte Carlo integration with importance sampling. 

\parsection{Grid Sampling}
For 2D image coordinate regression problems, \eg the case of regressing the center of the tracked target, $y \in \mathcal{Y} \subset \reals^2$ represents a location in the image. In this case, translational invariance is efficiently exploited by parametrizing $s_\theta(y,x) = f_\theta(x)(y)$, where $f_\theta$ is a Convolutional Neural Network (CNN). $s_\theta(y,x)$ is thus obtained by evaluating the output of the CNN at image coordinate $y$. Let $\{y^{(k)}\}_{k=1}^K \subset \mathcal{Y}$ be the set of uniform grid locations evaluated by the CNN $f_\theta(x)$ when applied to an image sample $x$. Further, let $A$ be the area of a single grid cell. The uniform grid sampling automatically provided by the CNN generates the following approximation of the loss \eqref{eq:kl-loss},
\begin{equation}
\label{eq:grid-loss}
L_i \! = \! \log \!\left(\!\!A\sum_{k=1}^{K} e^{s_\theta(y^{(k)},x_i)}\right) - A\!\sum_{k=1}^{K} s_\theta(y^{(k)},x_i) p(y^{(k)}|y_i).
\end{equation}
The final loss is then obtained by averaging $L_i$ over all samples $i$ in the mini-batch. 

\parsection{Monte Carlo Integration}
For the more general regression problems, grid sampling does not necessarily provide any computational benefits. On the contrary, it scales poorly to higher dimensions and may induce a sampling bias due to the rigid grid. In the more general case, we therefore adopt the Monte Carlo (MC) based sampling strategy proposed in~\cite{DCTD}. Specifically, we draw samples $y_i^{(k)} \sim q(y|y_i)$ from a proposal distribution $q(y|y_i)$ during training. The same samples are employed to approximate both integrals in \eqref{eq:kl-loss},
\begin{equation}
\label{eq:mc-loss}
L_i \!=\! \log \!\left(\!\!\frac{1}{K}\!\!\sum_{k=1}^{K} \frac{e^{s_\theta(y_i^{(k)}\!\!,x_i)}}{q(y_i^{(k)}|y_i)}\!\right) \!- \frac{1}{K}\!\sum_{k=1}^{K}\! s_\theta(y_i^{(k)}\!\!,x_i) \frac{p(y_i^{(k)}|y_i)}{q(y_i^{(k)}|y_i)}.
\end{equation}
To accurately approximate the original loss \eqref{eq:kl-loss}, the proposal distribution $q(y|y_i)$ should ideally cover the label distribution $p(y|y_i)$ as well as regions with high predicted density $p(y|x_i,\theta)$. In~\cite{DCTD} it was shown that a simple Gaussian mixture centered at the annotation $y_i$ sufficed for a variety of tasks, including bounding box regression.

The loss \eqref{eq:mc-loss} requires multiple evaluations of the network $s_\theta(y_i^{(k)},x_i)$. In practice, however, computer vision architectures popularly employ deep backbone feature extractors $\phi_\theta(x)$, such as ResNet~\cite{Resnet}, generating a powerful representation of the image. The output value $y$ can be fused at a late stage, such that $s_\theta(y,x) = f_\theta(y, \phi_\theta(x))$. This allows the computationally demanding feature extraction $\phi_\theta(x_i)$ to be shared among all samples $y_i^{(k)}$. Specifically for our purpose, such architectures have been successfully employed for bounding box regression in object detection and visual tracking problems~\cite{DiMP,ATOM,DCTD,IOUNet}. 

\section{Tracking Approach}
\label{sec:tracking}

We apply the general probabilistic regression formulation introduced in Section~\ref{sec:approach} for the challenging and diverse task of visual target tracking.

\subsection{Baseline Tracker: DiMP}
\label{sec:dimp}
We employ the recent state-of-the-art tracker DiMP~\cite{DiMP} as our baseline. As briefly discussed in Section~\ref{sec:confidence-tracking}, the DiMP model contains two output branches.

\parsection{Target Center Regression (TCR)}
The center regression branch aims to coarsely localize the target in the image by only regressing its center coordinate. This branch emphasizes robustness over accuracy. It consists of a linear convolutional output layer, who's weights $w_\theta$ are predicted by the network as an unrolled optimization process that minimizes an $L^2$-based discriminative learning loss. This allows the tracker to robustly differentiate the target object from similar objects in the background. The target center confidence at location $y^\text{tc} \in \reals^2$ in frame $x$ is predicted similarly to \eqref{eq:score-conv}, \ie $s_\theta^\text{tc}(y^\text{tc},x) = (w_\theta \conv \phi_\theta(x))(y^\text{tc})$, where $\phi_\theta$ is the backbone feature extractor. This branch is trained in a meta-learning setting, with a confidence-based objective \eqref{eq:confidence-loss} using Gaussian pseudo labels $a^\text{tc}$ and a robust $L^2$ loss,
\begin{equation}
\label{eq:dimp-robust-l2}
\ell(s,a) = \begin{dcases}
(s - a)^2 \,, & a > T \\
\max(0, s)^2 \,, & a \leq T
\end{dcases} \,.
\end{equation}
During tracking, the target center is regressed by densely computing confidence scores $s_\theta^\text{tc}(y^\text{tc},x)$ within a wide search region in the frame $x$. We refer to~\cite{DiMP} for details.

\parsection{Bounding Box Regression (BBR)}
The BBR branch adopts the target conditional IoU-Net-based~\cite{IOUNet} architecture proposed in~\cite{ATOM}. As discussed in Section~\ref{sec:confidence-tracking}, this branch predicts a confidence score $s_\theta^\text{bb}(y^\text{bb},x)$ for a given box \mbox{$y^\text{bb} \in \reals^4$}. It is trained using the bounding box IoU as pseudo label $a^\text{bb}(y^\text{bb},y^\text{bb}_i)$ and the standard $L^2$ loss $\ell$ in \eqref{eq:confidence-loss}. During tracking, the BBR branch is applied to fit an accurate bounding box to the target using gradient-based maximization of $s_\theta^\text{bb}(y^\text{bb},x)$ \wrt $y^\text{bb}$. We refer to~\cite{ATOM} for details.

\begin{figure}[!t]
	\centering%
	\newcommand{\widd}{0.25\columnwidth}%
	\newcommand{\implot}[1]{\includegraphics*[trim=68 70 140 50, width = \widd]{#1}}%
	\newcommand{\heatmapplot}[1]{\includegraphics*[trim=70 20 70 40, width = \widd]{#1}}%
	\newcommand{\probplotrow}[1]{%
		\implot{figures/cat/Figure_1_#1}%
		\heatmapplot{figures/cat/Figure_5_#1}%
		\heatmapplot{figures/cat/Figure_21_#1}%
		\heatmapplot{figures/cat/Figure_22_#1}}%
	\probplotrow{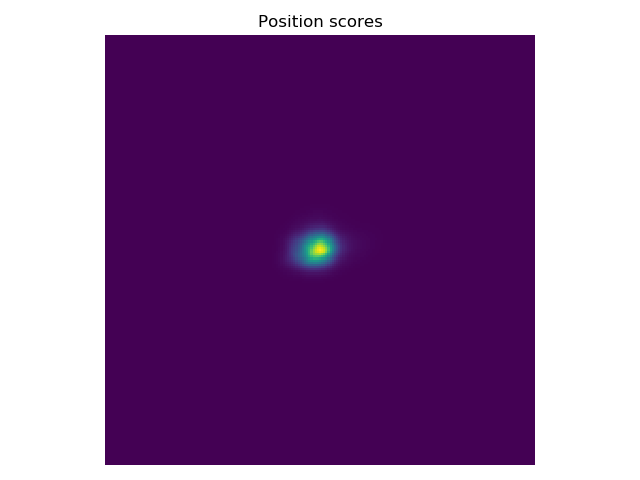}
	\probplotrow{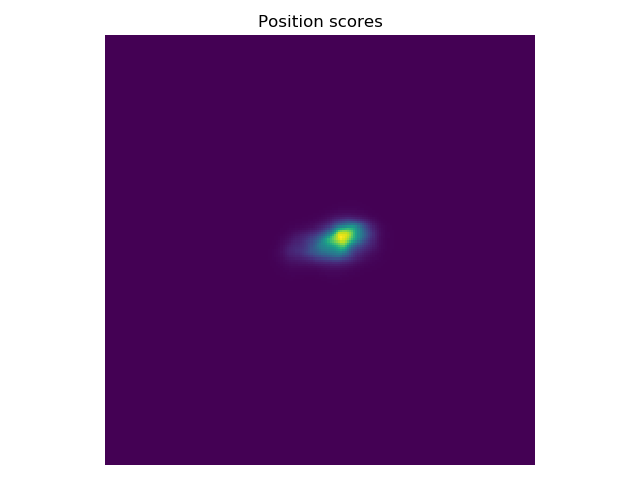}
	\probplotrow{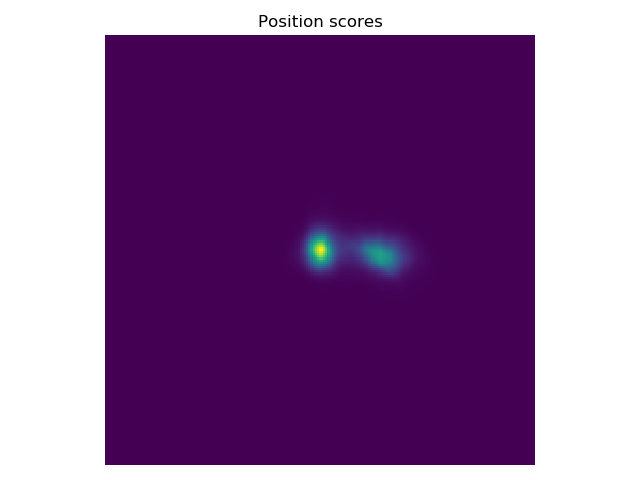}
	\resizebox{\columnwidth}{!}{
	\begin{tabular}{C{\widd}C{\widd}C{\widd}C{\widd}}
		Image & Target center regression & Bounding box center & Bounding box size
	\end{tabular}}%
	\caption{Visualization of the probability densities $p(y^\text{tc}|x,\theta)$ and $p(y^\text{bb}|x,\theta)$ predicted by the target center and bounding box regression branch respectively. The densities are centered at the predicted state (red box). The network captures uncertainties in the state, \eg larger variance or multiple modes, for challenging cases. More examples and discussion are provided in Appendix~\ref{sec:probvis}.}\vspace{-3mm}%
	\label{fig:probvis}%
\end{figure}

\subsection{Our Tracker: Probabilistic DiMP}
\label{sec:our-tracker}
We introduce a tracking approach with fully probabilistic output representations, obtained by integrating our regression formulation into both branches of the baseline DiMP. Example predicted densities are visualized in Figure~\ref{fig:probvis}.

\parsection{Target Center Regression}
We represent the predicted distribution of the target center coordinate $p(y^\text{tc}|x,\theta)$ by applying \eqref{eq:pred} to the network output $s_\theta^\text{tc}(y^\text{tc},x)$. Since this branch is fully convolutional, we approximate the KL-divergence loss \eqref{eq:kl-loss} for training using the grid sampling strategy \eqref{eq:grid-loss}. The conditional ground-truth density is set to a Gaussian $p(y^\text{tc}|y^\text{tc}_i)=\norm(y^\text{tc};y^\text{tc}_i,\sigma_\text{tc}^2)$ with the \emph{same} variance parameter $\sigma_\text{tc}^2$ used for the corresponding pseudo label function $a^\text{tc}$ in the baseline DiMP ($\frac{1}{4}$th of the target size). For the optimization module, which predicts the convolution weights $w_\theta$ for the center regression branch, we use the KL-divergence loss \eqref{eq:grid-loss} with an added $L^2$ regularization term. We modify the steepest descent based architecture~\cite{DiMP} to employ a second order Taylor expansion, since the original Gauss-Newton approximation is limited to least-squares objectives. Our approach benefits from the fact that the resulting objective \eqref{eq:grid-loss} is convex in $w_\theta$ for the linear predictor $s_\theta^\text{tc}(y^\text{tc},x) = (w_\theta \conv \phi_\theta(x))(y^\text{tc})$, and thanks to efficient analytic expressions of the gradient and Hessian. See Appendix~\ref{sec:tcr-details} for a detailed description of the optimizer module.

\parsection{Bounding Box Regression}
We use the same architecture $s_\theta^\text{bb}(y^\text{bb},x)$ as in~\cite{ATOM,DiMP} and apply it in our probabilistic formulation \eqref{eq:pred}. We follow the work of~\cite{DCTD}, which extended the same ATOM BBR module~\cite{ATOM} to the probabilistic setting using the negative log-likelihood loss \eqref{eq:nll} and an MC-based approximation. In this work, we further integrate the label distribution $p(y^\text{bb}|y^\text{bb}_i)$ to model the noise and uncertainty in the bounding box annotations, and minimize the KL-divergence \eqref{eq:kl-loss} using MC sampling \eqref{eq:mc-loss}. Specifically, we use an isotropic Gaussian distribution $p(y^\text{bb}|y^\text{bb}_i) = \norm(y^\text{bb};y^\text{bb}_i,\sigma_\text{bb}^2)$ and set $\sigma_\text{bb} = 0.05$. For a fair comparison, we use the same proposal distribution $q(y^\text{bb}|y^\text{bb}_i)=\frac{1}{2}\norm(y^\text{bb};y^\text{bb}_i,0.05^2)+\frac{1}{2}\norm(y^\text{bb};y^\text{bb}_i,0.5^2)$ and bounding box parametrization as in~\cite{DCTD}. 

\parsection{Details}
Our entire network is trained jointly end-to-end using the \emph{same} strategy and settings as for the original DiMP~\cite{DiMP}, by integrating it into the publicly available \mbox{PyTracking} framework~\cite{pytracking}. The training splits of the LaSOT~\cite{LaSOT}, GOT10k~\cite{GOT10k}, TrackingNet~\cite{TrackingNet} and COCO~\cite{COCO} are used, running 50 epochs with 1000 iterations each. We also preserve the tracking procedure and settings in DiMP, only performing minimal changes, which are forced by the probabilistic output representation provided by our model. Due to different scaling of the network output, we accordingly change the threshold for which the target is reported missing and the gradient step length used for bounding box regression. We refer to~\cite{DiMP,pytracking} for detailed description of training and inference settings. Our code is available at~~\cite{pytracking}.
\section{Experiments}
\label{sec:experiments}

We perform a detailed analysis of our approach along with comprehensive comparisons with state-of-the-art on seven datasets. The ResNet-18 and 50 versions of our tracker operates at about $40$ and $30$ FPS respectively.

\subsection{Comparison of Regression Models}
\label{sec:reg-analysis}

We first analyze the impact of different regression formulations for tracking. In each case, we train a separate version of the DiMP baseline, with the ResNet-18 backbone. We compare four different approaches. \textbf{L2:} Standard squared loss, used in the baseline DiMP for Bounding Box Regression (BBR). \textbf{R-L2:} The Robust L2 loss \eqref{eq:dimp-robust-l2}, employed in the baseline DiMP for the Target Center Regression (TCR) branch. \textbf{NLL:} The probabilistic Negative Log-Likelihood formulation \eqref{eq:nll} proposed in~\cite{DCTD}. \textbf{Ours:} Trained using the KL-divergence \eqref{eq:kl-loss} as described in Section~\ref{sec:our-tracker}. 

Following~\cite{DiMP}, we perform this analysis of our approach on the combined OTB-100~\cite{OTB2015}, UAV123~\cite{UAV123} and NFS~\cite{NfS} dataset, totaling 323 diverse videos, and report the average over 5 runs. We report the Overlap Precision $\text{OP}_{T}$, \ie percentage of frames with an IoU overlap larger than $T$, along with the main metric $\text{AUC}=\int_0^1 \text{OP}_{T}$ (see~\cite{OTB2015}).

\begin{table}[!t]
	\centering
	\resizebox{1.01\columnwidth}{!}{%
		\begin{tabular}{l@{~~}|@{~~}ccc@{~~}|@{~~}ccc@{~~}|@{~~}ccc@{~~}}
\toprule
&\multicolumn{3}{@{}c}{\textbf{Complete tracker}}&\multicolumn{3}{@{}c}{\textbf{Bounding Box reg.}}&\multicolumn{3}{@{}c}{\textbf{Target Center reg.}}\\
&\multicolumn{3}{@{}c}{Both BBR and TCR}&\multicolumn{3}{@{}c}{TCR formulation: \textbf{R-L2}}&\multicolumn{3}{@{}c}{BBR formulation: \textbf{Ours}}\\\midrule
Model& AUC & $\text{OP}_{0.50}$ & $\text{OP}_{0.75}$& AUC & $\text{OP}_{0.50}$ & $\text{OP}_{0.75}$& AUC & $\text{OP}_{0.50}$ & $\text{OP}_{0.75}$\\\midrule
L2& 63.1 & 78.0 & 50.2 & 63.8 & 79.2 & 50.6 & 64.8 & 80.9 & \textbf{54.1}\\
R-L2& 63.8 & 79.2 & 50.6 & 63.8 & 79.2 & 50.6 & \textbf{65.8} & \textbf{82.0} & \textbf{54.1}\\
NLL& 63.0 & 78.5 & 51.5 & 65.0 & 81.0 & 52.8 & 63.2 & 79.0 & 52.6\\
\textbf{Ours}& \textbf{65.5} & \textbf{81.6} & \textbf{54.1} & \textbf{65.8} & \textbf{82.0} & \textbf{54.1} & 65.5 & 81.6 & \textbf{54.1}\\\bottomrule
\end{tabular}

	}\vspace{1mm}%
	\caption{Analysis of four different regression models for tracking on the combined OTB100-NFS-UAV123 dataset. In the left section, the model is applied to both branches (BBR and TCR) of the network. In the center and right sections, we exclusively analyze their impact on BBR and TCR respectively. In the former case, R-L2 is always employed for TCR, while the right section uses Our approach for BBR in all cases. See text for details.}
	\label{tab:reg-analysis}%
	\vspace{-3mm}
\end{table}
\begin{figure}[b]
	\centering%
	\newcommand{\wid}{0.47\columnwidth}%
	\includegraphics[width = \wid]{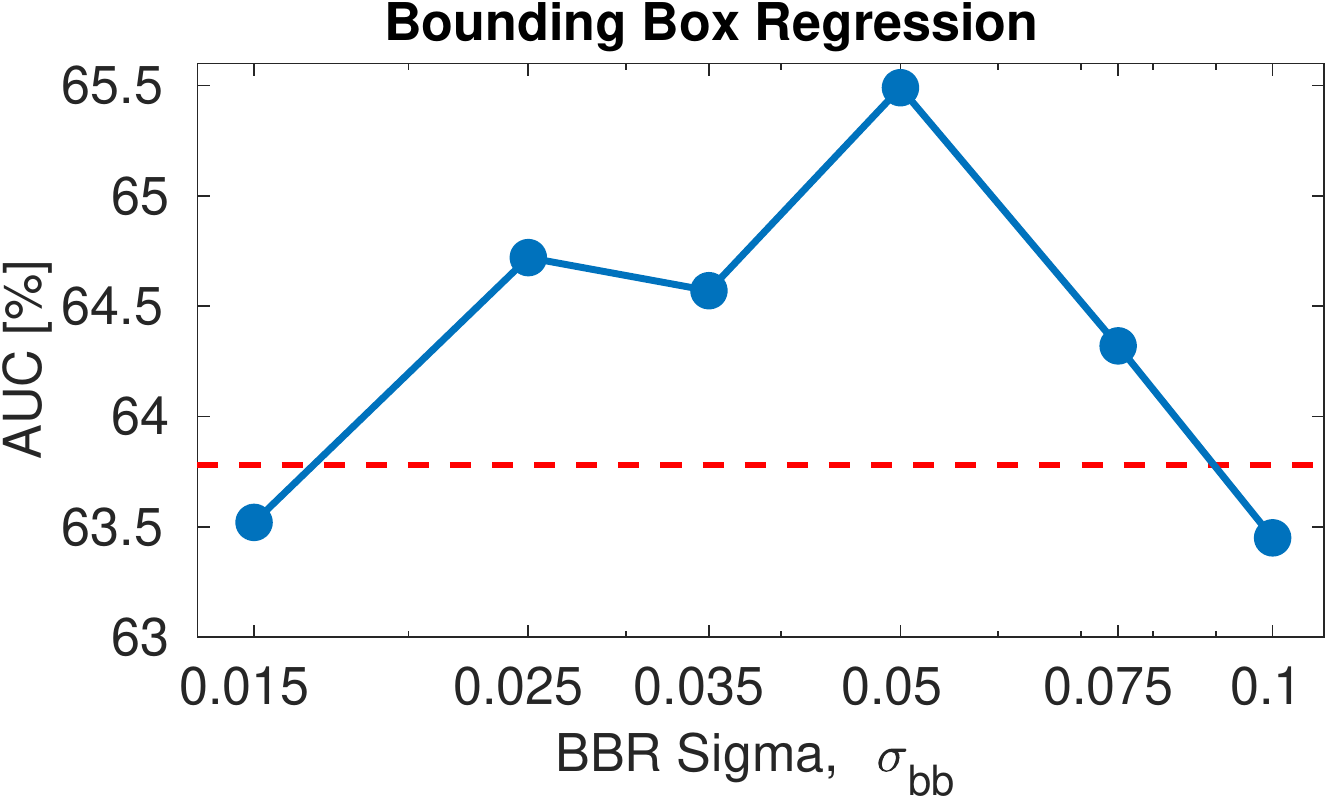}\hspace{3mm}%
	\includegraphics[width = \wid]{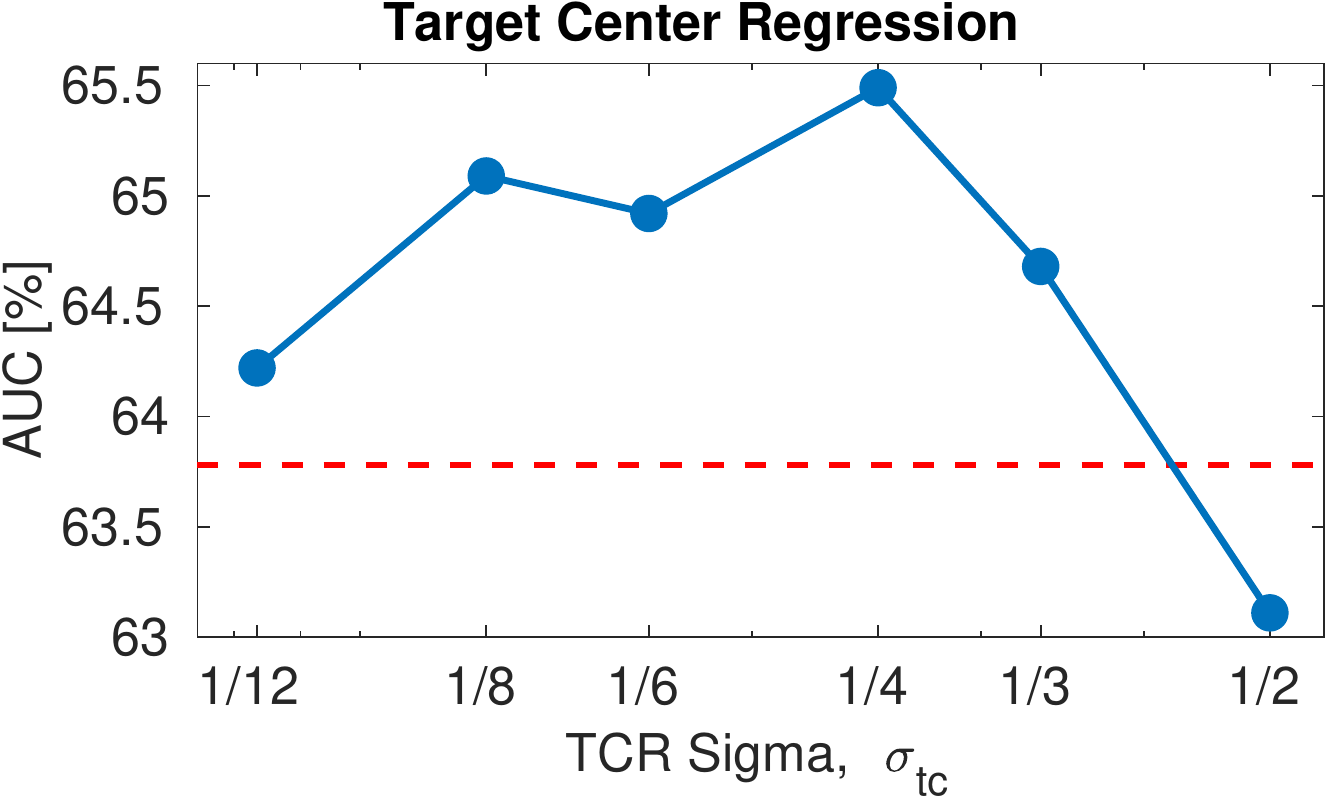}\vspace{-0.5mm}%
	\caption{The impact of modeling the label uncertainty $p(y|y_i)=\norm(y;y_i,\sigma^2)$ by varying the standard deviation $\sigma$ for bounding box regression (left) and target center regression (right). We show AUC on the combined OTB100-NFS-UAV123 datasets and display the baseline DiMP-18 result as a red dashed line.}%
	\label{fig:sigma}%
\end{figure}

\parsection{Complete Tracker}
We first analyze the performance when each regression model is applied to the entire tracker, \ie for \emph{both} the BBR and TCR branch. The results are reported in the left section of Table~\ref{tab:reg-analysis}. The R-L2, which corresponds to standard DiMP, improves $0.8\%$ AUC over the standard L2 loss. Our model outperforms the R-L2 baseline with a gain of $1.7\%$ in AUC. 

\parsection{Bounding Box Regression (BBR)}
We exclusively analyze the impact of each model for BBR by using the baseline DiMP R-L2 formulation for TCR in all cases. Thus, only the BBR branch is affected. Results are provided in the center section of Table~\ref{tab:reg-analysis}. The baseline~\cite{DiMP,ATOM}, employing L2 loss to predict IoU, achieves $63.8$ AUC. The NLL formulation~\cite{DCTD} gives a substantial $1.2\%$ gain. By modeling uncertainty in the annotation and using the KL loss \eqref{eq:kl-loss}, our approach achieves an additional improvement of $0.8\%$.

\parsection{Target Center Regression (TCR)}
Similarly, we compare the models for TCR by employing our approach for the BBR branch in all cases. The results, reported in the right section of Table~\ref{tab:reg-analysis}, show that probabilistic NLL is not suitable for TCR. This is likely due to the inherent ambiguity of the problem, which is not accounted for. By explicitly modeling label uncertainty $p(y|y_i)$, our formulation achieves a $2.3\%$ gain in AUC, further outperforming the standard L2 model. The R-L2 model, which is specifically designed for TCR~\cite{DiMP}, achieves a marginally better performance. However, note that this result is achieved when combined with \emph{our} formulation for BBR.

In conclusion, our probabilistic formulation outperforms the baseline DiMP, employing L2 and Robust L2 for TCR and BBR respectively, by a large $1.7\%$ margin in AUC, while achieving the best overall performance for all regression models. In the subsequent experiments, we employ our formulation for both the BBR and TCR branch.

\begin{figure}[t]
	\centering\vspace{-5mm}%
	\newcommand{\wid}{0.48\columnwidth}%
	\subfloat[LaSOT~\cite{LaSOT}\label{fig:lasot}]{\includegraphics[width = \wid]{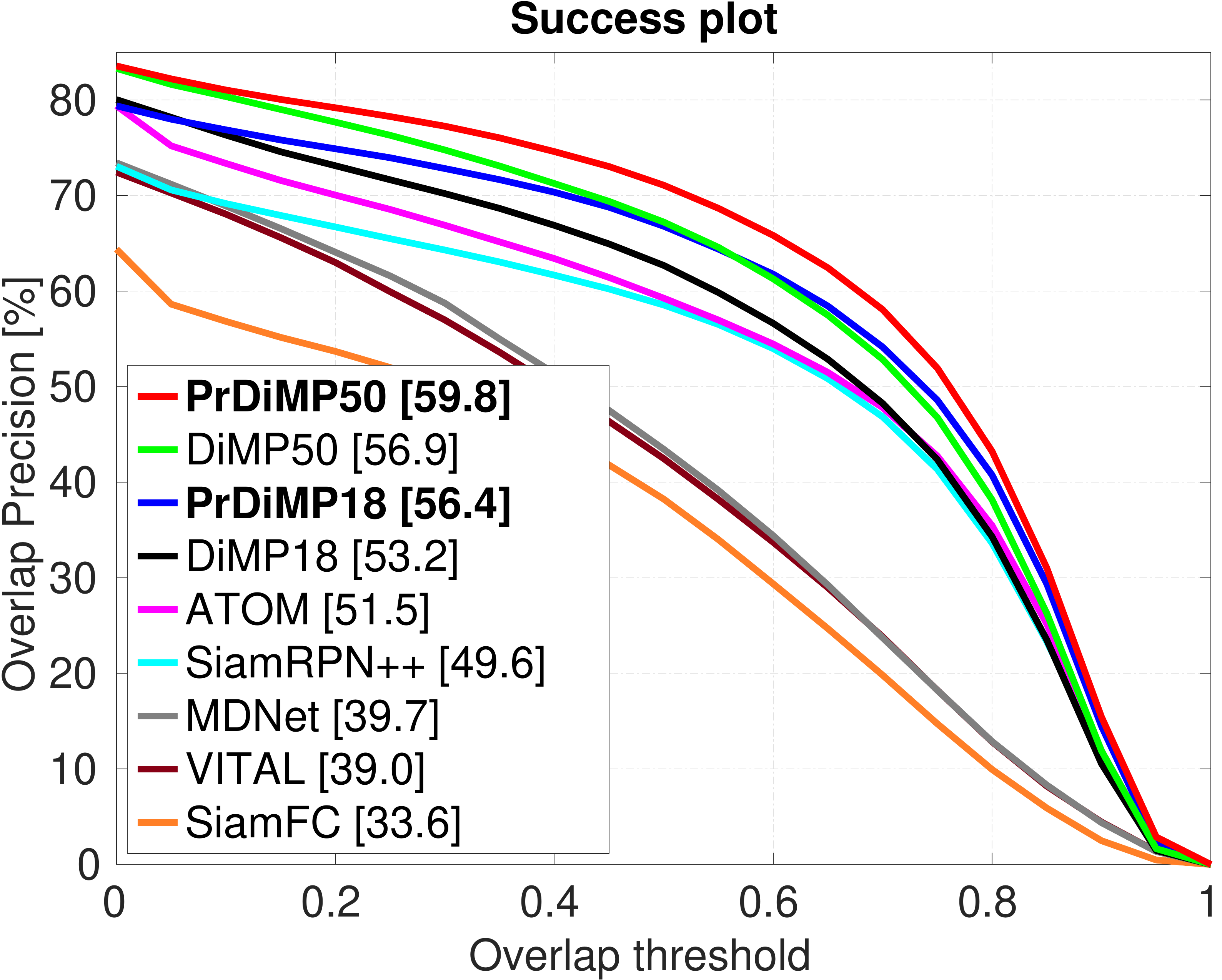}}~%
	\subfloat[UAV123~\cite{UAV123}\label{fig:uav}]{\includegraphics[width = \wid]{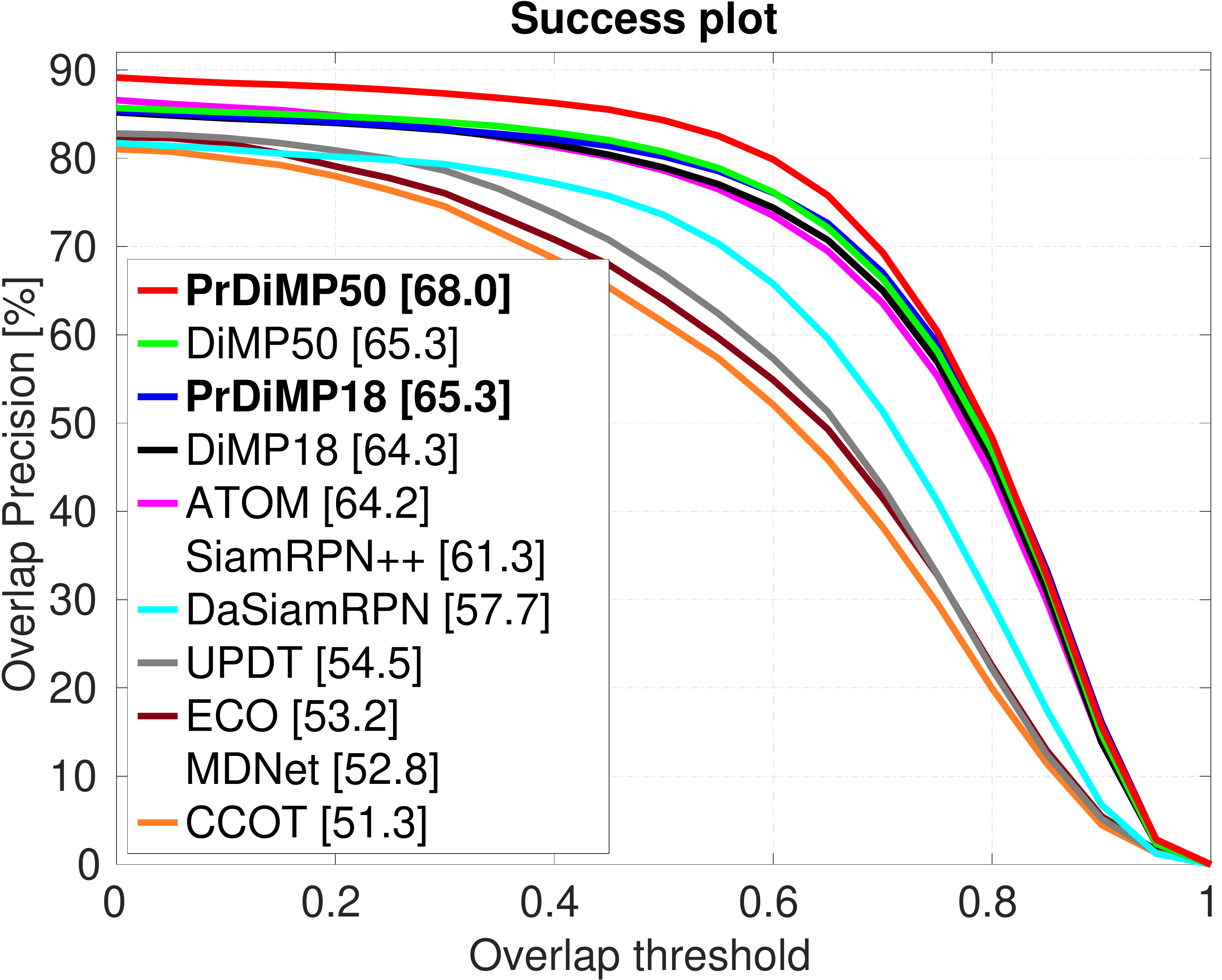}}%
	\caption{Success plots, showing $\text{OP}_T$, on LaSOT and UAV123, showing average over 5 runs for our method. Our approach outperforms DiMP by a large margin in AUC, shown in the legend.}%
	\label{fig:success}\vspace{-3mm}%
\end{figure}

\subsection{Analysis of Label Uncertainty}
\label{sec:label-uncertainty}

In this section, we further analyze the importance of our ability to model the uncertainty $p(y|y_i)$ in the annotation $y_i$. This is probed by investigating the impact of the standard deviation parameter $\sigma$ for our Gaussian model of the label noise $p(y|y_i) = \norm(y;y_i,\sigma^2)$ (see Section~\ref{sec:our-tracker}). This is individually performed for the bounding box ($\sigma_\text{bb}$) and target center ($\sigma_\text{tc}$) regression branch. We use the same experiment setup as described in the previous section, reporting the average of five runs over the combined OTB100-NFS-UAV123 dataset, and using the ResNet-18 backbone.

Figure~\ref{fig:sigma} shows the tracking performance in AUC when varying the standard deviations $\sigma_\text{bb}$ and $\sigma_\text{tc}$ over a wide range of values. The baseline DiMP-18 performance is shown in red for reference. A similar overall trend is observed in both cases. A too large standard deviation $\sigma$ leads to poor results, since the over-estimation in the label uncertainty forces the tracker to output too uncertain predictions. However, a small $\sigma$ causes overfitting and over-confident predictions, leading to sub-optimal performance. Properly modeling the label uncertainty is thus critical for visual tracking, due to the ambiguities in the regression tasks and annotation noise. We also note that by outperforming the baseline DiMP over a wide range of $\sigma_\text{bb}$ and $\sigma_\text{tc}$ values, our approach is not sensitive to specific settings.

\subsection{State-of-the-Art}
\label{sec:sota}

We compare our approach, termed \textbf{PrDiMP}, on seven tracking benchmarks. We evaluate two versions, PrDiMP18 and PrDiMP50, employing ResNet-18 and ResNet-50 respectively as backbone. The same settings and parameters are used for one-pass evaluation on all datasets. To ensure the significance in the results, we report the average over 5 runs for all datasets, unless the specific protocol requires otherwise. Additional results are provided in Appendix~\ref{sec:results}.

\parsection{LaSOT~\cite{LaSOT}}
We first compare on the large-scale LaSOT dataset. The test set contains 280 long videos (2500 frames in average), thus emphasizing the robustness of the tracker, along with its accuracy. The success plot in Figure~\ref{fig:lasot}, showing the overlap precision $\text{OP}_T$ as a function of the threshold $T$, is computed as the average over 5 runs. Trackers are ranked \wrt their AUC score, shown in the legend. Our approach outperforms the previous best tracker, DiMP, by a large margin of $3.2\%$ and $2.9\%$ in AUC when using ResNet-18 and ResNet-50 respectively. The improvement in  $\text{OP}_T$ is most prominent for $T>0.3$, demonstrating superior accuracy achieved by our bounding box regression.

\parsection{TrackingNet~\cite{TrackingNet}}
This is a large-scale tracking dataset with high variety in terms of classes and scenarios. The test set contains over 500 videos without publicly available ground-truth. The results, shown in Table~\ref{tab:trackingnet}, are obtained through an online evaluation server. Both our versions outperform all previous approaches in the main Success (AUC) metric by a significant margin. With the same ResNet-50 backbone, our approach achieves a gain of $2.5\%$ and $1.8\%$ in AUC over SiamRPN++~\cite{SiamRPN++} and DiMP~\cite{DiMP} respectively.

\parsection{VOT2018~\cite{VOT2018}}
Next, we evaluate on the 2018 edition of the Visual Object Tracking challenge. We compare with the top methods in the challenge~\cite{VOT2018}, as well as more recent methods. The dataset contains 60 videos. Trackers are restarted at failure by the evaluation system. The performance is then decomposed into accuracy and robustness, defined using IoU overlap and failure rate respectively. The main EAO metric takes both these aspects into account. The results, computed over 15 repetitions as specified in the protocol, are shown in Table~\ref{tab:vot}. Our PrDiMP50 achieves the best overall performance, with the highest accuracy and competitive robustness compared to previous methods.

\begin{table}[!t]
	\centering
	\resizebox{1.01\columnwidth}{!}{%
		\begin{tabular}{l@{~}c@{~~}c@{~~}c@{~~}c@{~~}c@{~~}c@{~~}c@{~~}c@{~~}c@{~~}c@{~~}}
	\toprule
	&SiamFC&MDNet&UPDT&DaSiam-&ATOM&SiamRPN++&DiMP18&DiMP50&\textbf{PrDiMP18}&\textbf{PrDiMP50}\\
	&\cite{SiameseFC}&\cite{MDNet}&\cite{BhatECCV2018}&RPN \cite{DaSiamRPN}&\cite{ATOM}&\cite{SiamRPN++}&\cite{DiMP}&\cite{DiMP}&&\\\midrule
	Precision &53.3&56.5&55.7&59.1&64.8&\textbf{\textcolor{blue}{69.4}}&66.6&68.7&69.1&\textbf{\textcolor{red}{70.4}}\\
	Norm.\ Prec.\ &66.6&70.5&70.2&73.3&77.1&80.0&78.5&80.1&\textbf{\textcolor{blue}{80.3}}&\textbf{\textcolor{red}{81.6}}\\
	Success (AUC) &57.1&60.6&61.1&63.8&70.3&73.3&72.3&74.0&\textbf{\textcolor{blue}{75.0}}&\textbf{\textcolor{red}{75.8}}\\\bottomrule
\end{tabular}

	}\vspace{1mm}%
	\caption{Results on the TrackingNet~\cite{TrackingNet} test set in terms of precision, normalized precision, and success (AUC). Both our PrDiMP versions outperform previous methods by a significant margin.
	}
	\label{tab:trackingnet}%
	\vspace{-2mm}
\end{table}
\begin{table}[!b]
	\centering\vspace{-5mm}
	\resizebox{1.01\columnwidth}{!}{%
		\begin{tabular}{l@{~}c@{~~}c@{~~}c@{~~}c@{~~}c@{~~}c@{~~}c@{~~}c@{~~}c@{~~}c@{~~}c@{~~}}
	\toprule
	&RCO&UPDT&DaSiam-&MFT&LADCF&ATOM&SiamRPN++&DiMP18&DiMP50&\textbf{PrDiMP18}&\textbf{PrDiMP50}\\
	&\cite{VOT2018}&\cite{BhatECCV2018}&RPN \cite{DaSiamRPN}&\cite{VOT2018}&\cite{LADCF}&\cite{ATOM}&\cite{SiamRPN++}&\cite{DiMP}&\cite{DiMP}&&\\\midrule
	EAO&0.376&0.378&0.383&0.385&0.389&0.401&0.414&0.402&\textbf{\textcolor{blue}{0.440}}&0.385&\textbf{\textcolor{red}{0.442}}\\
	Robustness&0.155&0.184&0.276&\textbf{\textcolor{red}{0.140}}&0.159&0.204&0.234&0.182&\textbf{\textcolor{blue}{0.153}}&0.217&0.165\\
	Accuracy&0.507&0.536&0.586&0.505&0.503&0.590&0.600&0.594&0.597&\textbf{\textcolor{blue}{0.607}}&\textbf{\textcolor{red}{0.618}}\\\bottomrule
\end{tabular}

	}\vspace{1mm}%
	\caption{Results on the VOT2018 challenge dataset~\cite{VOT2018} in terms of expected average overlap (EAO), robustness and accuracy. 
	}
	\label{tab:vot}%
\end{table}

\parsection{GOT10k~\cite{GOT10k}}
This dataset contains $10,000$ sequences for training and $180$ for testing. We follow the defined protocol~\cite{GOT10k} and \emph{only} train on the specified GOT10k training set for this experiments, while keeping all other settings the same. By having no overlap in object classes between training and testing, GOT10k also benchmarks the generalizability of the tracker to novel objects. The results in Table~\ref{tab:got10k}, obtained through a evaluation server, are reported in terms of $\text{SR}_T$ and AO, which are equivalent to $\text{OP}_T$ and AUC~\cite{GOT10k}. Compared to DiMP, our method achieve large gains of $3.3\%$ and $2.3\%$ in terms of the main AO metric, when using ResNet-18 and -50 respectively.

\parsection{UAV123~\cite{UAV123}}
This challenging dataset, containing 123 videos, is designed to benchmark trackers for UAV applications. It features small objects, fast motions, and distractor objects. The results are shown in Figure~\ref{fig:uav}, where $\text{OP}_T$ is plotted over IoU thresholds $T$ and AUC is shown in the legend. ATOM~\cite{ATOM} and DiMP50 obtain $64.2\%$ and $65.3\%$ respectively. Our PrDiMP50 achieves a remarkable $68.0\%$, surpassing previous methods by a significant margin.

\parsection{OTB-100~\cite{OTB2015}}
For reference, we report results on the OTB-100 dataset. While this dataset has served an important role in the development of trackers since its release, it is known to have become highly saturated over recent years, as shown in Table~\ref{tab:nfs_uav_otb}. Still, our approach performs similarly to the top correlation filter methods, such as UPDT~\cite{BhatECCV2018}, while on par with the end-to-end trained SiamRPN++.

\parsection{NFS~\cite{NfS}}
Lastly, we report results on the 30 FPS version of the Need for Speed (NFS) dataset, containing fast motions and challenging distractors. As shown in Table~\ref{tab:nfs_uav_otb}, our approach achieves a substantial improvement over the previous state-of-the-art on this dataset.

\begin{table}[!t]
	\centering
	\resizebox{1.01\columnwidth}{!}{%
		\begin{tabular}{l@{~~}c@{~~}c@{~~}c@{~~}c@{~~}c@{~~}c@{~~}c@{~~}c@{~~}c@{~~}c@{~~}c@{~~}}
	\toprule
	&CF2&ECO&CCOT&GOTURN&SiamFC&SiamFCv2&ATOM&DiMP18&DiMP50&\textbf{PrDiMP18}&\textbf{PrDiMP50}\\
	&\cite{HCF_ICCV15}&\cite{DanelljanCVPR2017}&\cite{DanelljanECCV2016}&\cite{Held2016gotrun}&\cite{SiameseFC}&\cite{Valmadre2017cvpr}&\cite{ATOM}&\cite{DiMP}&\cite{DiMP}&&\\\midrule
	SR$_{0.50}$ &29.7&30.9&32.8&37.5&35.3&40.4&63.4&67.2&\textbf{\textcolor{blue}{71.7}}&71.5&\textbf{\textcolor{red}{73.8}}\\
	SR$_{0.75}$ &8.8&11.1&10.7&12.4&9.8&14.4&40.2&44.6&49.2&\textbf{\textcolor{blue}{50.4}}&\textbf{\textcolor{red}{54.3}}\\
	AO &31.5&31.6&32.5&34.7&34.8&37.4&55.6&57.9&61.1&\textbf{\textcolor{blue}{61.2}}&\textbf{\textcolor{red}{63.4}}\\\bottomrule
\end{tabular}

	}\vspace{1mm}%
	\caption{State-of-the-art comparison on the GOT10k test set~\cite{GOT10k}. The metrics average overlap (AO) and success rate ($\text{SR}_T$) are equivalent to AUC and $\text{OP}_T$ respectively (described in Section~\ref{sec:reg-analysis}). The evaluated methods are trained only on the GOT10k training set, and must therefore generalize to novel classes.
	}
	\label{tab:got10k}%
	\vspace{-1mm}
\end{table}
\begin{table}[!t]
	\centering
	\resizebox{1.01\columnwidth}{!}{%
		\begin{tabular}{l@{~}c@{~~}c@{~~}c@{~~}c@{~~}c@{~~}c@{~~}c@{~~}c@{~~}c@{~~}c@{~~}c@{~~}}
	\toprule
	&DaSiam-&CCOT&MDNet&ECO&ATOM&SiamRPN++&UPDT&DiMP18&DiMP50&\textbf{PrDiMP18}&\textbf{PrDiMP50}\\
	&RPN \cite{DaSiamRPN}&\cite{ATOM}&\cite{DanelljanECCV2016}& \cite{MDNet}&\cite{DanelljanCVPR2017}&\cite{SiamRPN++}&\cite{BhatECCV2018}&\cite{DiMP}&\cite{DiMP}&&\\\midrule
	OTB-100&65.8&68.2&67.8&69.1&66.9&\textbf{\textcolor{blue}{69.6}}&\textbf{\textcolor{red}{70.2}}&66.0&68.4&68.0&\textbf{\textcolor{blue}{69.6}}\\
	NFS&-&48.8&42.2&46.6&58.4&-&53.7&61.0&62.0&\textbf{\textcolor{blue}{63.3}}&\textbf{\textcolor{red}{63.5}}\\\bottomrule
\end{tabular}

	}\vspace{1mm}%
	\caption{Comparison with state-of-the-art on the OTB-100~\cite{OTB2015} and NFS~\cite{NfS} datasets in terms of overall AUC score. The average value over 5 runs is reported for our approach.}
	\label{tab:nfs_uav_otb}%
	\vspace{-2mm}
\end{table}
\section{Conclusions}
\label{sec:conclusions}
We propose a probabilistic regression formulation, where the network is trained to predict the conditional density $p(y|x,\theta)$ of the output $y$ given the input $x$. The density is parametrized by the architecture itself, allowing the representation of highly flexible distributions. The network is trained by minimizing the KL-divergence to the label distribution $p(y|y_i)$, which is introduced to model annotation noise and task ambiguities. When applied for the tracking task, our approach outperforms the baseline DiMP \cite{DiMP} and sets an new state-of-the-art on six datasets.

\noindent\textbf{Acknowledgments:}
This work was partly supported by the ETH Z\"urich Fund (OK), a Huawei Technologies Oy (Finland) project, an Amazon AWS grant, and Nvidia.

{\small
\bibliographystyle{ieee_fullname}
\bibliography{references}
}

\clearpage

\setcounter{section}{0}
\renewcommand{\thesection}{\Alph{section}}

\begin{center}
	\textbf{\Large Appendix}
\end{center}

This Appendix provides detailed derivations and results.
Appendix~\ref{sec:klloss} derive the employed loss in \eqref{eq:kl-loss} from the KL divergence. We then provide details about the target center regression module in Appendix~\ref{sec:tcr-details}. Lastly, we report more detailed results in Appendix~\ref{sec:results} and provide additional visualizations of the predicted distributions in Appendix~\ref{sec:probvis}.

\section{Derivation of KL Divergence Loss}
\label{sec:klloss}
Here, we derive the loss \eqref{eq:kl-loss} from the KL divergence between the predicted distribution $p(y|x_i,\theta)$ and the ground-truth density $p(y|y_i)$. Starting from the definition of the KL divergence and inserting our formulation \eqref{eq:pred}, we achieve,
\begin{align}
\label{eq:kl-loss-deriv}
& \KL(p(\cdot|y_i),p(\cdot|x_i,\theta)) = \int p(y|y_i) \log \frac{p(y|y_i)}{p(y|x_i,\theta)} \diff y \nonumber \\ 
& = \int p(y|y_i) \log \frac{p(y|y_i)}{e^{s_\theta(y,x_i)} / Z_\theta(x_i)} \diff y \nonumber \\
& = \int p(y|y_i) \left( \log p(y|y_i) - \log e^{s_\theta(y,x_i)} + \log Z_\theta(x_i) \right) \diff y \nonumber \\
& = \int p(y|y_i) \log p(y|y_i) \diff y \nonumber \\
& \qquad\qquad + \log Z_\theta(x_i) - \int p(y|y_i) s_\theta(y,x) \diff y \nonumber \\
&\sim \log \left(\int e^{s_\theta(y,x_i)} \diff y\right) - \int s_\theta(y,x_i) p(y|y_i)  \diff y \,.
\end{align}
In the last row, we have discarded the first term (the negative entropy of $p(y|y_i)$) and substituted the definition of the partition function $Z_\theta(x_i)$ from \eqref{eq:pred}.

\section{Target Center Regression Module}
\label{sec:tcr-details}

In this section, we give a detailed description and derivation of the optimization module employed for target center regression in our PrDiMP tracker. 
The goal of the optimizer module is to predict the weights $w_\theta$ of the target center regression component,
\begin{equation}
\label{eq:tcr}
s_\theta^\text{tc}(y^\text{tc},x) = (w_\theta \conv \phi_\theta(x))(y^\text{tc}) \,.
\end{equation}
Here, $x$ is an input image, $y^\text{tc} \in \reals^2$ is an image coordinate and $\phi_\theta$ is the backbone feature extractor (see Section~\ref{sec:our-tracker}). The weights $w_\theta$ are learned from a set of training (support) images $\{z_j\}_{j=1}^n$ and corresponding target bounding box annotations $\{\tilde{y}_j^\text{bb}\}_{j=1}^n$. As in the baseline DiMP~\cite{DiMP}, these images are sampled from an interval within each sequence during training. During tracking, the images $z_j$ are obtain by performing augmentations on the first frame, and by gradual update of the memory.

In DiMP, the optimizer module is derived by applying the steepest descent algorithm to a least-squares objective. In our case however, the employed loss function does not admit a least-squares formulation. In this work, we therefore replace the Gauss-Newton approximation with the quadratic Newton approximation in order to compute the step-length necessary for the steepest descent algorithm. We derive closed form solutions of all operations, ensuring simple integration of the optimizer module as a series of deep neural network layers.

Similarly to our offline training objective, we let the optimizer module minimize the KL-divergence based learning loss \eqref{eq:kl-loss}. We also add an $L^2$ regularization term to benefit generalization to unseen frames. The loss is thus formulated as follows,
\begin{equation}
\label{eq:loss-main}
L(w_\theta) = \sum_{j=1}^{n} \gamma_j L_\text{CE}(\tilde{z}_j \conv w_\theta;p_j) + \frac{\lambda}{2} \|w_\theta\|^2 \,.
\end{equation}
The non-negative scalars $\lambda$ and $\gamma_j$ control the impact of the regularization term and sample $z_j$ respectively. We also make the following definitions for convenience,
\begin{subequations}
\label{eq:definitions}
\renewcommand{\arraystretch}{1.5}
\arraycolsep 0.2em
\begin{equationarray}{rl@{\hspace{3mm}}l@{\hspace{9mm}}}
\tilde{z}_j &= \phi_\theta(z_j) & \text{Extracted image features.} \label{eq:features} \\
p_j^{(k)}\!\!\! &= p(y^{\text{tc},(k)}|y_j^{\text{tc}}) & \text{Label density at location $k$.} \label{eq:gt-prob} \\
s_j^{(k)}\!\!\!  &= (\tilde{z}_j \!\conv\! w_\theta)(y^{\text{tc},(k)}) & \text{Target scores at location $k$.} \label{eq:scores} \\
\hat{p}_j^{(k)}\!\!\! &= \frac{\exp(s_j^{(k)})}{\sum_{l=1}^{K} \exp(s_j^{(l)})} & \text{Spatial SoftMax of } s_j \text{.} \label{eq:softmax}
\end{equationarray}
\end{subequations}
Note that we use superscript $k \in \{1,\ldots,K\}$ to denote spatial grid location $y^{\text{tc},(k)} \in \reals^2$. In the following, the quantities in \eqref{eq:gt-prob}-\eqref{eq:softmax} are either seen as vectors in $\reals^K$ or 2D-maps $\reals^{H\times W}$ (with $K=HW$), as made clear from the context. 

In \eqref{eq:loss-main}, the per-sample loss $L_\text{CE}$ is the grid approximation \eqref{eq:grid-loss} of the original KL-divergence objective. Without loss of generality, we may assume $A=1$, obtaining
\begin{align}
	\label{eq:ce-loss}
	L_\text{CE}(s;p) &= \log \left( \sum_{k=1}^{K} e^{s^{(k)}}\right) - \sum_{k=1}^{K} s^{(k)} p^{(k)} \nonumber \\
	&= \log \left(\mathbf{1}\tp e^s\right) - p\tp s\,.
\end{align}
Here, $\mathbf{1}\tp=[1, \ldots, 1]$ denotes a vector of ones. Note that the grid approximation thus corresponds to the SoftMax-Cross Entropy loss, commonly employed for classification. 

\newcommand{\init}{\psi_\theta^\text{init}}
\newcommand{\assign}{\leftarrow}
\newcommand{\algcomment}[1]{\hfill\textit{#1}}
\begin{algorithm}[t]
	\caption{Target Model Prediction $\psi_\theta$ for Center Reg.}
	\begin{algorithmic}[1]
		\Require Training (support) images $\{z_j\}_{j=1}^n$
		\Require Corresponding box annotations $\{\tilde{y}_j^\text{bb}\}_{j=1}^n$\vspace{1mm}
		\State $\tilde{z}_j = \phi_\theta(z_j),\quad j=1,\ldots,n$ 	\algcomment{Extract features}
		\State $w^{(0)}_\theta \assign \init\left(\{(z_j,\tilde{y}_j^\text{bb})\}_{j=1}^n\right)$ \algcomment{Initialize weights \cite{DiMP}}
		\For{$i = 0, \ldots, N_\text{iter} - 1$} \algcomment{Optimizer module loop}
		\State $s_j \assign \tilde{z}_j \conv w^{(i)}_\theta$ \algcomment{Using \eqref{eq:features}}
		\State $\hat{p}_j \assign \text{SoftMax}(s_j)$ \algcomment{Using \eqref{eq:softmax}}
		\State $g \assign \nabla L(w^{(i)}_\theta)$ \algcomment{Using \eqref{eq:loss-main-grad}}
		\State $\mathlarger{\alpha^{(i)} = \frac{g\tp g}{g\tp H(w_\theta^{(i)}) g}}$ \algcomment{Using \eqref{eq:hess-vec} for the denom.}
		\State $w_\theta^{(i+1)} \assign w_\theta^{(i)} - \alpha^{(i)} g$ \algcomment{Update weights}
		\EndFor
	\end{algorithmic}
	\label{alg:model-predictor}
\end{algorithm}

To derive the optimization module, we adopt the steepest descent formulation \cite{DiMP}, but employ the Newton approximation discussed above. This results in the following optimization strategy,
\begin{subequations}
\label{eq:sd}
\begin{align}
w_\theta^{(i+1)} &= w_\theta^{(i)} - \alpha^{(i)} \nabla L(w_\theta^{(i)}) \\ 
\alpha^{(i)} &= \frac{\nabla L(w_\theta^{(i)})\tp \nabla L(w_\theta^{(i)})}{\nabla L(w_\theta^{(i)})\tp H(w_\theta^{(i)}) \nabla L(w_\theta^{(i)})} \,.
\end{align}
\end{subequations}
Here, $\nabla L(w_\theta^{(i)})$ and $H(w_\theta^{(i)})$ is the gradient and Hessian of $L$ \eqref{eq:loss-main}, evaluated at the current estimate $w_\theta^{(i)}$ of the weights. To implement \eqref{eq:sd}, we efficiently compute these quantities by deriving closed-form expressions of both. 

First, we can first easily compute the gradient and hessian of \eqref{eq:ce-loss} \wrt $s$ as,
\begin{subequations}
	\label{eq:ce-grad-hess} 
\begin{align}
	\nabla_s L_\text{CE}(s;p) &= \hat{p} - p \label{eq:ce-grad} \\
	\frac{\partial^2}{\partial s^2} L_\text{CE}(s;p) &= \diag(\hat{p}) - \hat{p}\hat{p}\tp \label{eq:ce-hess}
\end{align}
\end{subequations}
Here, $\hat{p}$ is the SoftMax of $s$ as defined in \eqref{eq:softmax}. By applying \eqref{eq:ce-grad} together with the chain rule, we can compute the gradient of \eqref{eq:loss-main} as,
\begin{align}
\label{eq:loss-main-grad}
\nabla L(w_\theta) &= \sum_{j=1}^{n} \gamma_j \left[\frac{\partial s_j}{\partial w_\theta}\right]\tp \nabla_s L_\text{CE}(s_j;p_j) + \lambda w_\theta \nonumber \\
&= \sum_{j=1}^{n} \gamma_j \tpconvmat{\tilde{z}_j}(\hat{p}_j - p_j) + \lambda w_\theta \,.
\end{align}
We denote the transpose of the linear convolution operator $w \mapsto \tilde{z} \conv w$ as $\tpconvmat{\tilde{z}}$, which corresponds to the transpose of the Jacobian $\frac{\partial s_j}{\partial w_\theta}=\convmat{\tilde{z}}$. By another differentiation \wrt $w_\theta$, using \eqref{eq:ce-hess}, the chain rule and the linearity of $s_j$ in $w_\theta$, we obtain the Hessian of \eqref{eq:loss-main} as
\begin{align}
\label{eq:loss-main-hess}
H(&w_\theta) =  \frac{\partial^2}{\partial w_\theta^2} L(w_\theta) \nonumber \\
&= \sum_{j=1}^{n} \gamma_j \left[\frac{\partial s_j}{\partial w_\theta}\right]\tp \left[\frac{\partial}{\partial^2 s} L_\text{CE}(s_j;p_j)\right] \left[\frac{\partial s_j}{\partial w_\theta}\right] + \lambda I \nonumber \\
&= \sum_{j=1}^{n} \gamma_j \tpconvmat{\tilde{z}_j}(\diag(\hat{p}_j) - \hat{p}_j\hat{p}_j\tp)\convmat{\tilde{z}_j} + \lambda I \,.
\end{align}
Here, $I$ denotes the identity matrix. We further obtain a simple expression of the denominator of the step-length in \eqref{eq:sd} by evaluating the product,
\begin{align}
\label{eq:hess-vec}
g\tp &H(w_\theta)g = \\ 
&= \sum_{j=1}^{n} \gamma_j (\tilde{z}_j \conv g)\tp (\diag(\hat{p}_j) - \hat{p}_j\hat{p}_j\tp) (\tilde{z}_j \conv g) + \lambda g\tp g \nonumber \\ 
&= \sum_{j=1}^{n} \gamma_j v_j\tp \big(\hat{p}_j \cdot (v_j - \hat{p}_j\tp v_j)\big) + \lambda g\tp g \;,\quad v_j = \tilde{z}_j \conv g \,. \nonumber
\end{align}
Here, $\cdot$ denotes element-wise multiplication. We summarize the full optimization module in Algorithm~\ref{alg:model-predictor}.

\section{Detailed Results}
\label{sec:results}

We provide more detailed results from the state-of-the-art comparison performed in Section~\ref{sec:sota}.

\begin{figure}[b]
	\centering%
	\newcommand{\wid}{0.48\textwidth}%
	\includegraphics*[trim = 0 0 0 0, width = \wid]{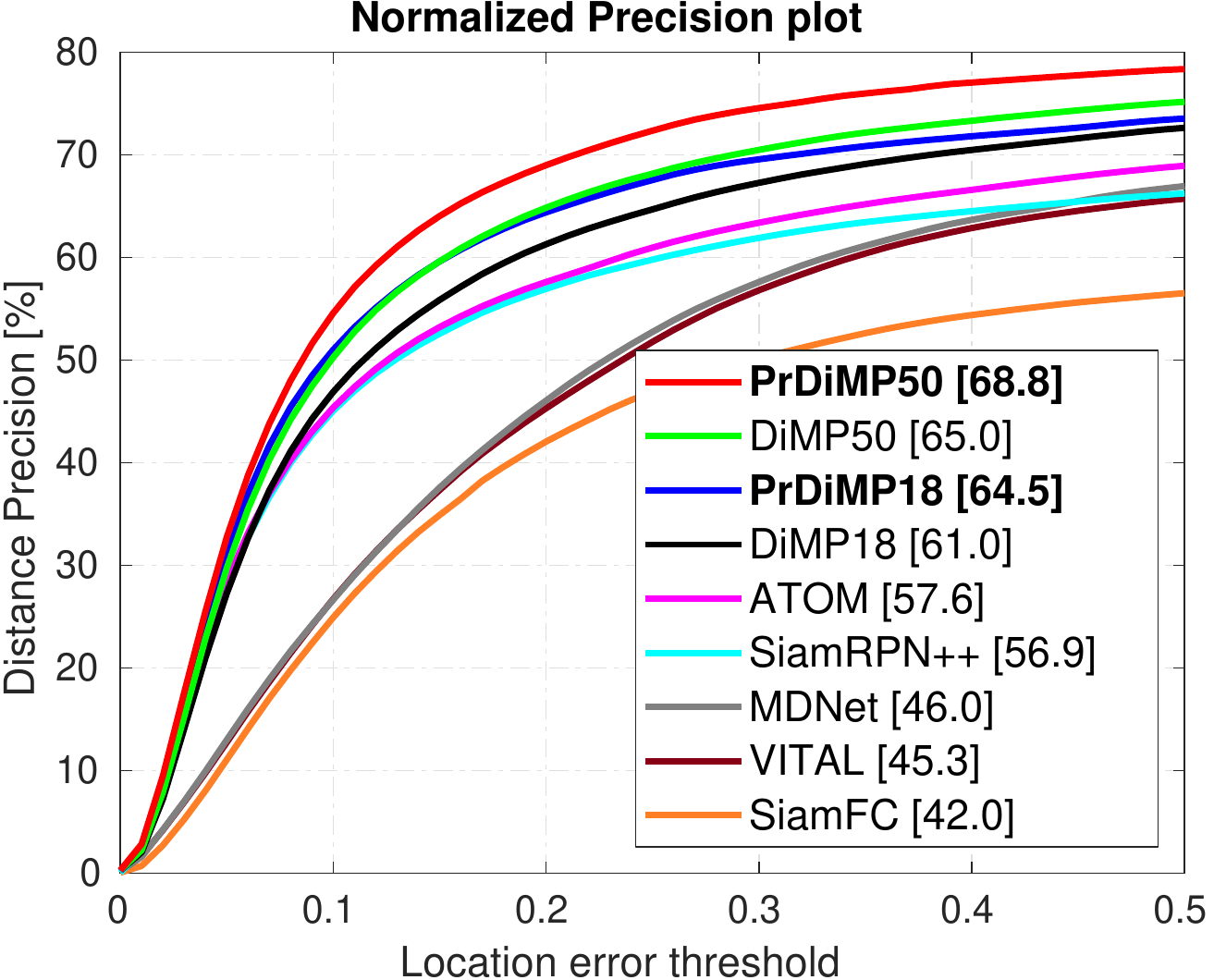}%
	\caption{Normalized precision plot on the LaSOT dataset~\cite{LaSOT}. The average normalized precision is shown in the legend. Our approach outperforms previous trackers by a large margin.
	}%
	\label{fig:lasot-prec}%
\end{figure}
\begin{figure*}[t]
	\newcommand{\wid}{0.33\textwidth}%
	\centering\vspace{-5mm}%
	\subfloat[GOT10k~\cite{GOT10k}\label{fig:got}]{\includegraphics[trim = 0 0 0 0,width = \wid]{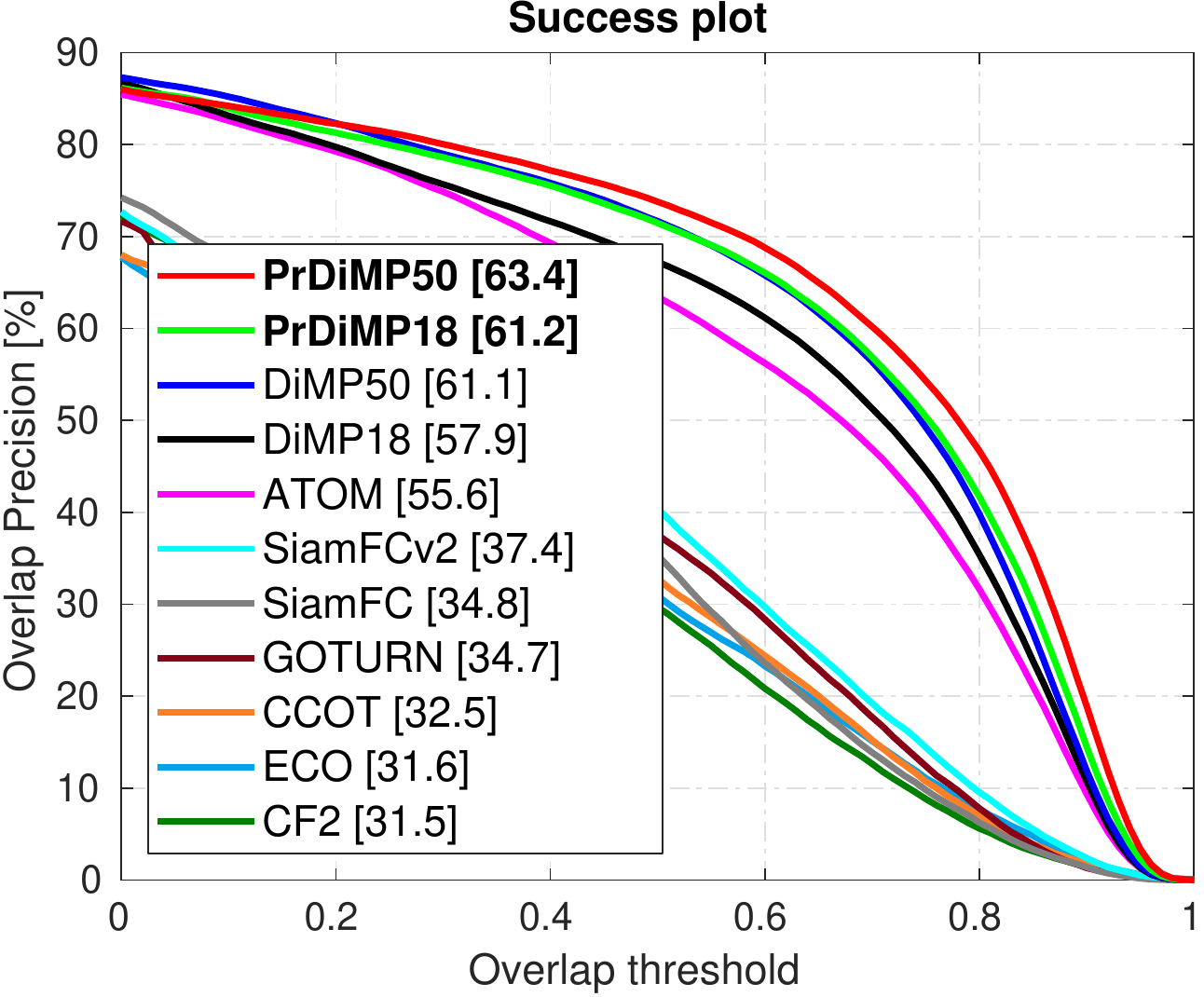}}%
	\subfloat[OTB-100~\cite{OTB2015}\label{fig:otb}]{\includegraphics[trim = 0 0 0 0,width = \wid]{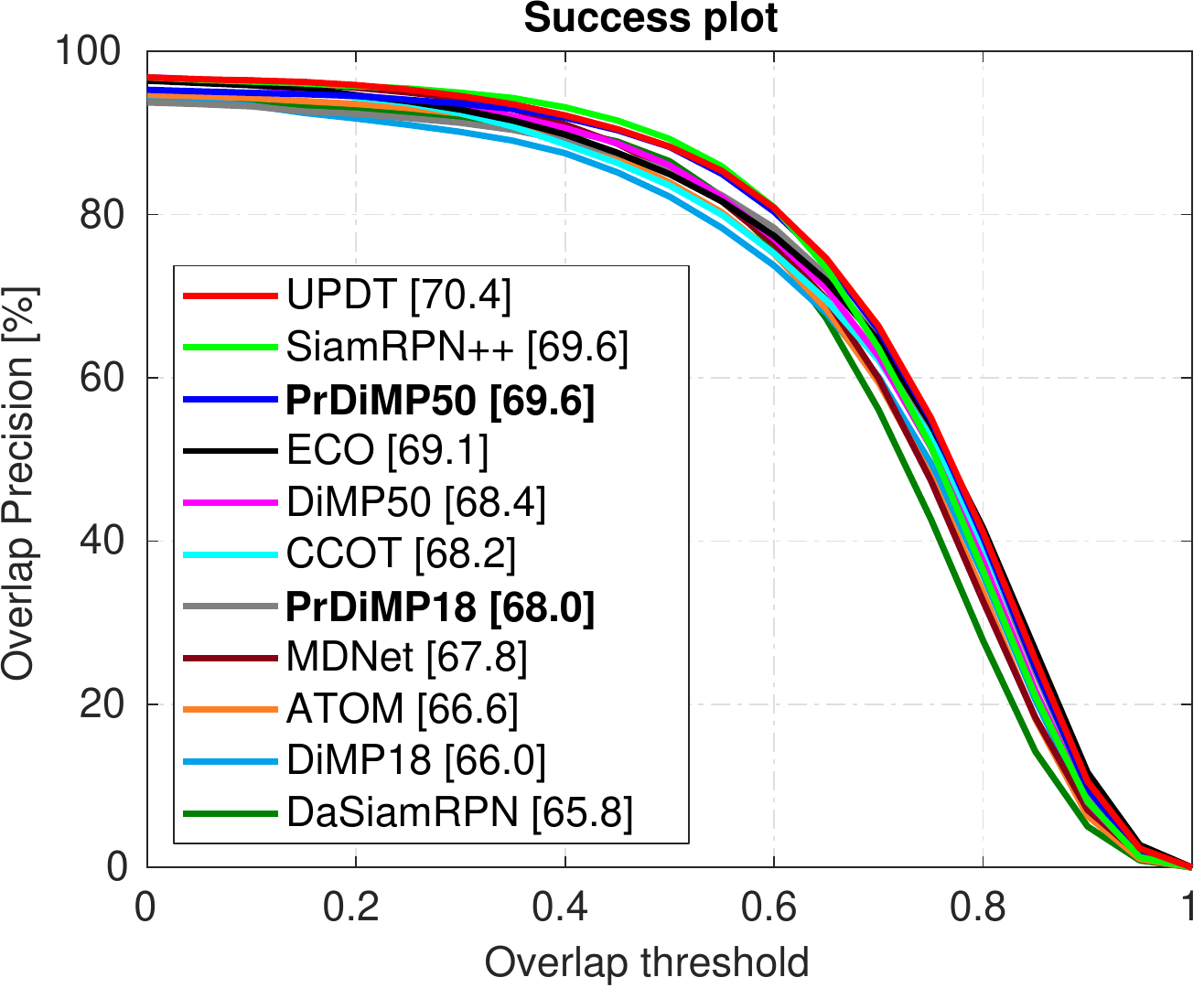}}%
	\subfloat[NFS~\cite{NfS}\label{fig:nfs}]{\includegraphics[trim = 0 0 0 0, width = \wid]{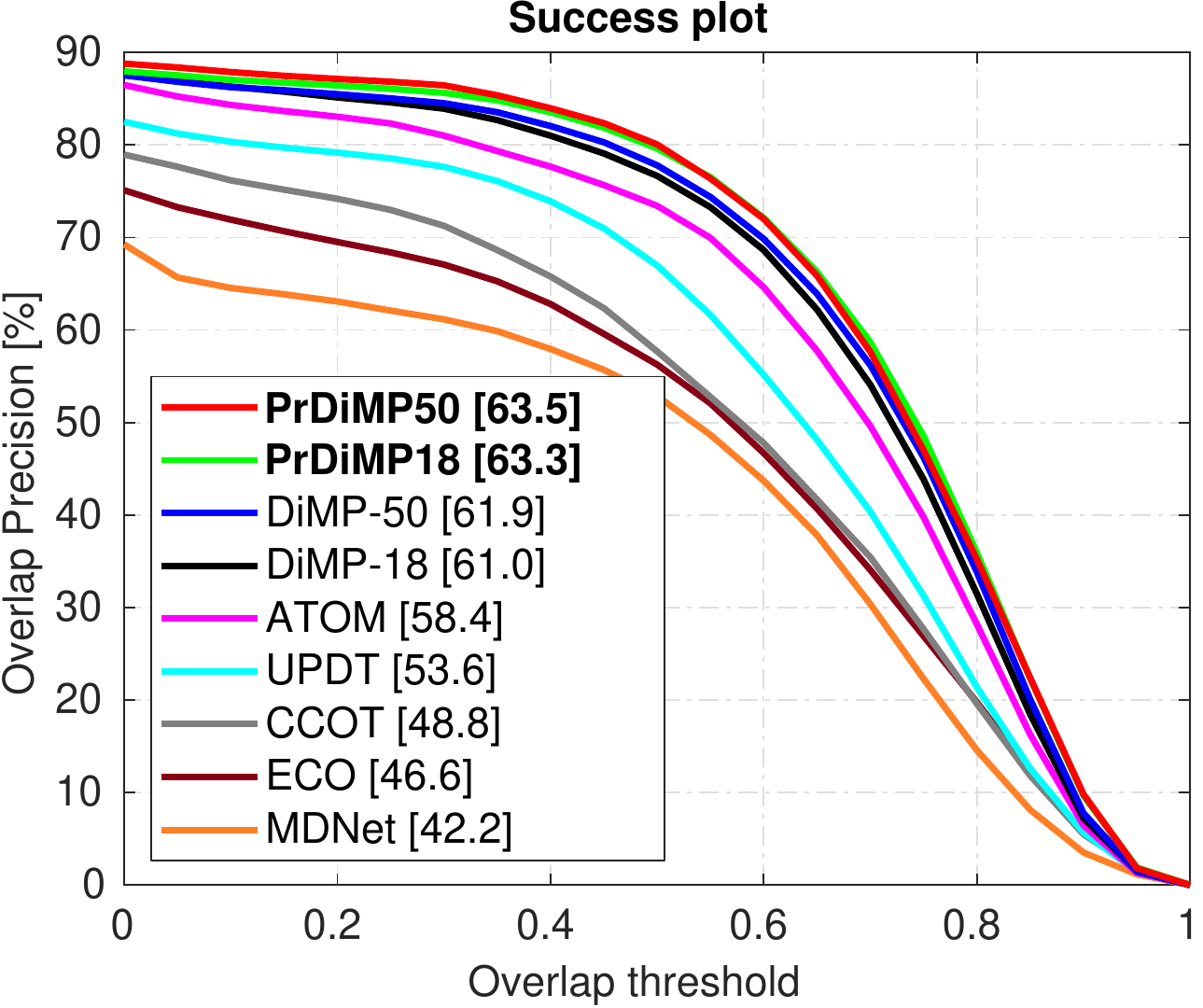}}\vspace{1mm}%
	\caption{Success plots on the GOT10k~\cite{GOT10k} (a), OTB-100~\cite{OTB2015} (b), and NFS~\cite{NfS} (c) datasets, showing the percentage of frames with a ground-truth IoU overlap larger than a threshold. The area-under-the-curve (AUC) metric for each tracker is shown in the legend. Our approach outperforms previous approaches by a significant margin on the challenging GOT10k and NFS datasets, while performing similarly to the top trackers on the largely saturated OTB-100 dataset.}\vspace{-2mm}%
	\label{fig:success_plots}
\end{figure*}

\subsection{LaSOT}
In addition to the success plot shown in Section~\ref{sec:sota}, we here provide the normalized precision plot over the LaSOT~\cite{LaSOT} test set, containing 280 videos. The normalized precision score $\text{NPr}_D$ is computed as the percentage of frames where the normalized distance (relative to the target size) between the predicted and ground-truth target center location is less than a threshold $D$. $\text{NPr}_D$ is plotted over a range of thresholds $D \in [0,0.5]$. The trackers are ranked using the area under this curve, which is shown in the legend (see~\cite{LaSOT} for details). The normalized precision plot is shown in Figure~\ref{fig:lasot-prec}. For the ResNet-18 and 50 versions of our PrDiMP, we report the average result over 5 runs. We compare with state-of-the-art trackers DiMP~\cite{DiMP}, ATOM~\cite{ATOM}, SiamRPN++~\cite{SiamRPN++}, MDNet~\cite{MDNet}, VITAL~\cite{VITAL}, and SiamFC~\cite{SiameseFC}. Our approach outperforms previous state-of-the-art by a large margin. Compared to the ResNet-50 based SiamRPN++~\cite{SiamRPN++} and DiMP-50~\cite{DiMP}, our PrDiMP50 version achieves absolute gains of $11.9\%$ and $3.8\%$ respectively. As demonstrated by the plot, this gain is obtained by an improvement in both accuracy (small thresholds) and robustness (large thresholds).

\subsection{GOT10k}
The success plot over the 180 videos in the GOT10k~\cite{GOT10k} test set is shown in Figure~\ref{fig:got}. It displays the overlap precision $\text{OP}_T$ as a function of the IoU threshold $T$. Overlap precision $\text{OP}_T$ itself is defined as the percentage of frames with a ground-truth IoU overlap larger than a threshold $T$. The final area-under-the-curve (AUC) score is shown in the legend. We compare our approach with trackers with available results: DiMP~\cite{DiMP}, ATOM~\cite{ATOM}, SiamFCv2 (CFNet)~\cite{Valmadre2017cvpr}, SiamFC~\cite{SiameseFC}, GOTRUN~\cite{Held2016gotrun}, CCOT~\cite{DanelljanECCV2016}, ECO~\cite{DanelljanCVPR2017}, and CF2 (HCF)~\cite{HCF_ICCV15}. Our PrDiMP versions outperform previous methods, in particular for large overlap thresholds. This demonstrates the superior accuracy of our probabilistic bounding box regression formulation. 

\subsection{OTB-100}
We provide the success plot over the 100 videos in the OTB dataset~\cite{OTB2015} in Figure~\ref{fig:otb}. We compare with state-of-the-art trackers UPDT~\cite{BhatECCV2018}, SiamRPN++~\cite{SiamRPN++}, ECO~\cite{DanelljanCVPR2017}, DiMP~\cite{DiMP}, CCOT~\cite{DanelljanECCV2016}, MDNet~\cite{MDNet}, ATOM~\cite{ATOM}, and DaSiamRPN~\cite{DaSiamRPN}. Despite the highly saturated nature of this dataset, our tracker performs among the top methods, providing a significant gain over the baseline DiMP.

\subsection{NFS}
Figure~\ref{fig:nfs} show the success plot over the 30 FPS version of the challenging NFS dataset~\cite{NfS}. We compare with top trackers with available results:  DiMP~\cite{DiMP}, ATOM~\cite{ATOM}, UPDT~\cite{BhatECCV2018}, CCOT~\cite{DanelljanECCV2016}, ECO~\cite{DanelljanCVPR2017}, and MDNet~\cite{MDNet}. In this case, both our ResNet-18 and 50 versions achieve similar results, significantly outperforming the previous state-of-the-art DiMP-50.

\begin{figure*}[b]
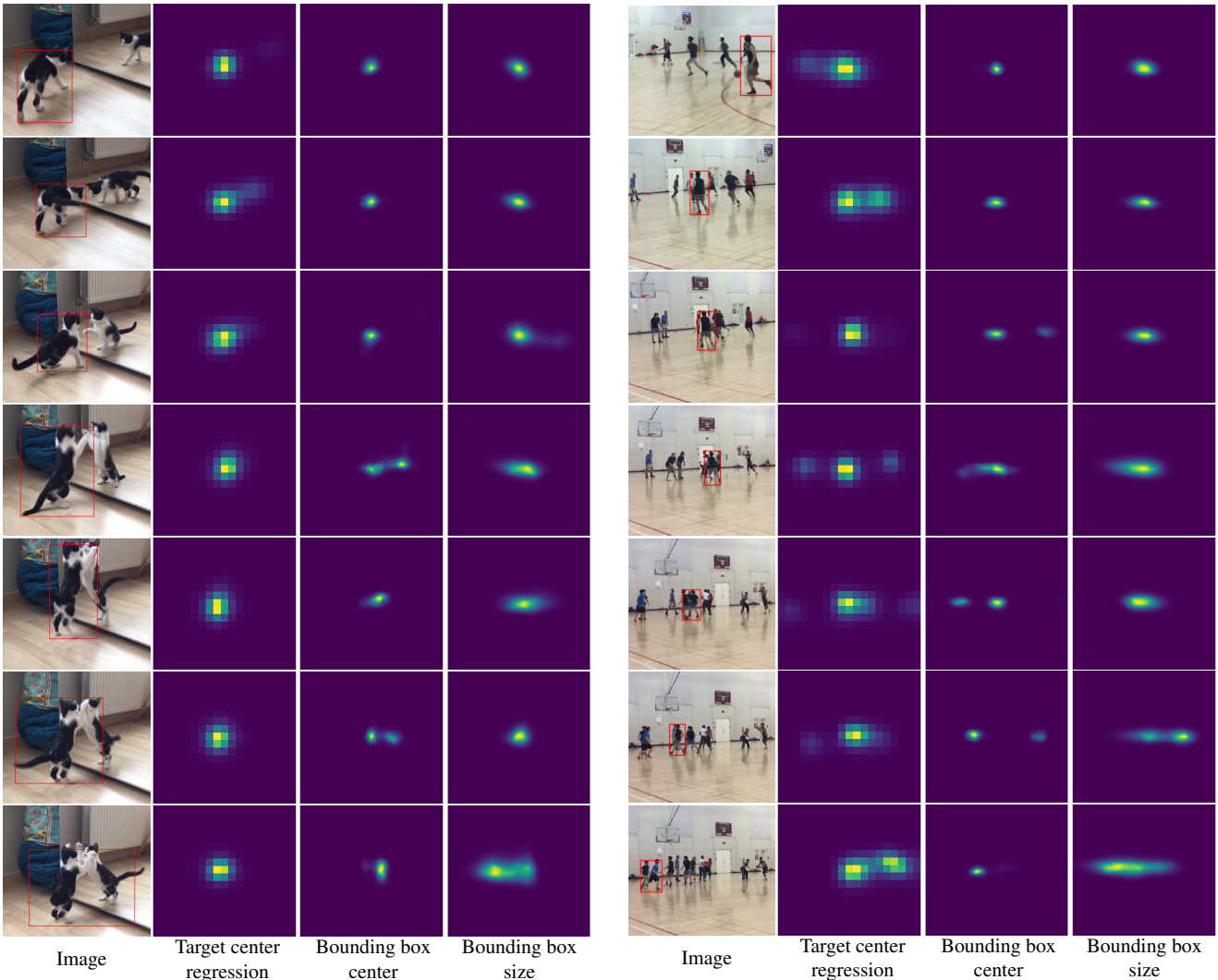

	\centering\vspace{-2mm}%
	\newcommand{\widd}{0.25\columnwidth}%
	\newcommand{\implot}[1]{\includegraphics*[trim=68 70 140 50, width = \widd]{#1}}%
	\newcommand{\heatmapplot}[1]{\includegraphics*[trim=70 20 70 40, width = \widd]{#1}}%
	\newcommand{\probplotrow}[2]{%
		\implot{figures/#1/Figure_1_#2}%
		\heatmapplot{figures/#1/Figure_5_#2}%
		\heatmapplot{figures/#1/Figure_21_#2}%
		\heatmapplot{figures/#1/Figure_22_#2}}%
	\begin{minipage}{0.485\textwidth}
		\probplotrow{cat}{0}
		\probplotrow{cat}{2}
		\probplotrow{cat}{3}
		\probplotrow{cat}{4}
		\probplotrow{cat}{5}
		\probplotrow{cat}{6}
		\probplotrow{cat}{7}
		\resizebox{\columnwidth}{!}{
			\begin{tabular}{C{\widd}C{\widd}C{\widd}C{\widd}}
				Image & Target center regression & Bounding box center & Bounding box size
		\end{tabular}}
	\end{minipage}\hfill%
	\begin{minipage}{0.485\textwidth}
		\renewcommand{\implot}[1]{\includegraphics*[trim=120 100 176 100, width = \widd]{#1}}%
		\probplotrow{basketball}{1}
		\probplotrow{basketball}{8}
		\probplotrow{basketball}{10}
		\probplotrow{basketball}{15}
		\probplotrow{basketball}{18}
		\probplotrow{basketball}{19}
		\probplotrow{basketball}{20}
		\resizebox{\columnwidth}{!}{
			\begin{tabular}{C{\widd}C{\widd}C{\widd}C{\widd}}
				Image & Target center regression & Bounding box center & Bounding box size
		\end{tabular}}
	\end{minipage}%
	\caption{Visualization of the probability densities $p(y^\text{tc}|x,\theta)$ and $p(y^\text{bb}|x,\theta)$ predicted by the target center and bounding box regression branch respectively. We illustrate the output for example frames in two highly challenging sequences, where capturing uncertainty is important. Refer to the text for details.}%
	\label{fig:probvis-suppl}%
\end{figure*}

\section{Visualization of Predicted Distributions}
\label{sec:probvis}

In this section, we provide additional visualizations of the predicted probability distributions for target center and bounding box regression branches in our tracker. We visualize the output as in Figure~\ref{fig:probvis}. Several example frames for two challenging sequences are shown in Figure~\ref{fig:probvis-suppl}. As during standard tracking, the predicted distribution $p(y^\text{tc}|x,\theta)$ for the target center regression (second column) is computed by applying the fully convolutional center regression branch on the search region centered at the previous target location.

To visualize the probability distribution $p(y^\text{bb}|x,\theta)$ predicted by the bounding box regression branch, we evaluate the density in a grid. Note that the bounding box $y^\text{bb} \in \reals^4$ is 4-dimensional, and we therefore cannot visualize the full distribution as a 2-dimensional heatmap. We therefore plot two \emph{slices} of this density, showing the variation in the bounding box location and size as follows. Our bounding box is parametrized as $y^\text{bb} = (c_x/w_0, c_y/h_0, \log w, \log h)$, where $(c_x,c_y)$ is the center position, $(w,h)$ is the size (width and height), and $(w_0,h_0)$ is a constant reference size. The latter is set to the current estimate of the target size. The distribution of the bounding box center (third column in Figure~\ref{fig:probvis-suppl}) is obtained by predicting the density value $p(y^\text{bb}|x,\theta) \sim \exp(s_\theta^\text{bb}(y^\text{bb},x))$ in a dense grid of center coordinates $(c_x,c_y)$, while keeping the size $(w,h)$ constant at the current target estimate. To visualize the distribution over the bounding box size (fourth column), we conversely evaluate $p(y^\text{bb}|x,\theta) \sim \exp(s_\theta^\text{bb}(y^\text{bb},x))$ in a dense grid of log-size coordinates $(\log w, \log h)$, while keeping the box centered at the current target estimate $(c_x,c_y)$. 

In Figure~\ref{fig:probvis-suppl}, the target center density is visualized in the range $\pm 2.5\sqrt{wh}$ relative to the previous target location. The bounding box center density is plotted within the range $c_x \pm w$ and $c_y \pm h$. The distribution over the bounding box size is plotted from $1/3$ to $3$ times the estimated target size $(w,h)$, \ie $\pm \log 3$. Outputs for two challenging sequences are visualized in Figure~\ref{fig:probvis-suppl}. The left part shows a cat and its mirror image. Due to their similarity and proximity, it is in many frames difficult to predict the exact bounding box of the cat. In these cases, the predicted distribution captures this uncertainty. For example, in the fourth row, two clear modes are predicted, which correspond to aligning the box with the real cat or with the right edge of the reflection. Moreover, the last row shows a failure case, where the box briefly expands. However, note that the size probability distribution (right column) is highly uncertain. This information could thus be used to indicate low reliability of the estimated bounding box size.

The right part of Figure~\ref{fig:probvis-suppl} depicts a challenging sequence with multiple distractors. The target center regression, which has a wider view of the scene, captures the presence of distractors in uncertain cases. In the last row, the tracker briefly fails by jumping to a distractor object. However, there is a strong secondary mode in the target center regression distribution that indicates the true target. Thus, the estimated distribution accurately captures the uncertainty in this ambiguous case. Moreover, in row 4-7, the target object is small with many nearby similar-looking objects. This makes bounding box regression extremely hard, even for a human. Our network can predict flexible distributions reflecting meaningful uncertainties for both bounding box position and size, when encountered with these difficulties.

\end{document}